\documentclass{article}
\usepackage{aaai19}

\usepackage{times}
\usepackage{soul}
\usepackage[utf8]{inputenc}
\usepackage[small]{caption}
\usepackage{amsmath,amssymb}
\usepackage[pdftex]{graphicx} 

\usepackage{subfigure}

\usepackage{bm}
\usepackage{url}
\usepackage{fancyhdr}
\usepackage{mathtools}
\usepackage{mathrsfs}
\newcommand{\equationname}{Equation~}
\renewcommand{\figurename}{Figure~}
\renewcommand{\tablename}{Table~}

\title{Partially Non-Recurrent Controllers for Memory-Augmented Neural Networks}
\author{
\begin{minipage}{0.49\textwidth}
\centering
Naoya Taguchi\thanks{Work done while the author was at the University of Tokyo} \\
{\normalfont 
    DeNA Co., Ltd., Shibuya Hikarie 2-21-1 \\ Shibuya, Shibuya-ku Tokyo, Japan \\
    naoya.taguchi@dena.com
}
\end{minipage}
\begin{minipage}{0.52\textwidth}
\centering
Yoshimasa Tsuruoka \\
{\normalfont 
    The University of Tokyo, 3-7-1 \\ Hongo, Bunkyo-ku, Tokyo, Japan \\
    tsuruoka@logos.t.u-tokyo.ac.jp
}
\end{minipage}
}

\author{
  Naoya Taguchi\thanks{Work done while the author was at the University of Tokyo}\footnotemark[2] \and Yoshimasa Tsuruoka\footnotemark[3]
  \\ 
  \footnotemark[2]DeNA Co., Ltd., Shibuya Hikarie 2-21-1 Shibuya, Shibuya-ku Tokyo, Japan \\
  \footnotemark[3]The University of Tokyo, 3-7-1 Hongo, Bunkyo-ku, Tokyo, Japan \\
  \footnotemark[2]astt.hwhw@gmail.com, \footnotemark[3]tsuruoka@logos.t.u-tokyo.ac.jp
}

\author{
  Naoya Taguchi\thanks{Work done while the author was at the University of Tokyo}\footnotemark[2] \and Yoshimasa Tsuruoka\footnotemark[3]
  \\\\
  \footnotemark[2]DeNA Co., Ltd., Shibuya Hikarie 2-21-1 Shibuya, Shibuya-ku Tokyo, Japan \\
  naoya.taguchi@dena.com \\\\
  \footnotemark[3]The University of Tokyo, 3-7-1 Hongo, Bunkyo-ku, Tokyo, Japan \\
  tsuruoka@logos.t.u-tokyo.ac.jp
}

\begin{document}

\maketitle

\begin{abstract}
    Memory-Augmented Neural Networks (MANNs) are a class of neural networks equipped with an external memory, and are reported to be effective for tasks requiring a large long-term memory and its selective use. The core module of a MANN is called a controller, which is usually implemented as a recurrent neural network (RNN) (e.g., LSTM) to enable the use of contextual information in controlling the other modules. However, such an RNN-based controller often allows a MANN to directly solve the given task by using the (small) internal memory of the controller, and prevents  the MANN from making the best use of the external memory, thereby resulting in a suboptimally trained model. To address this problem, we present a novel type of RNN-based controller that is partially non-recurrent and avoids the direct use of its internal memory for solving the task, while keeping the ability of using contextual information in controlling the other modules. %Without increasing the number of parameters, our proposed RNN-based controller avoids the direct use of the internal memory of the controller for solving tasks. 
Our empirical experiments using Neural Turing Machines and Differentiable Neural Computers on the Toy and bAbI tasks demonstrate that the proposed controllers give substantially better results than standard RNN-based controllers.
%\textbf{Our empirical experiments on the Toy and bAbI tasks demonstrate that the proposed controllers often give substantially better results than standard RNN-based controllers.}

    %Memory-Augmented Neural Networks (MANNs) are a novel type of Recurrent Neural Networks (RNNs) with external memory, and they perform well on tasks which require large long-term memory and its selective use. The core module of MANNs is the controller, which is usually implemented using RNNs to allow the models to utilize context information for controlling the other modules. However, these RNN-based controllers also allow the models to use the memory in the controller to solve tasks directly, which makes them away from appropriate learning in some cases. In this paper, we improve the performance of MANNs by introducing a novel type of RNN-based controller which does not suffer from this problem. Without increasing the number of parameters, our proposed RNN-based controller avoids the direct use of the internal memory of the controller for solving tasks. Experimental results on the Toy tasks and the bAbI task show that the proposed RNN-based controllers achieve better performance on both of the tasks than other types of controllers.

\end{abstract}

\section{Introduction}
\label{introduction}
%Applications such as Natural Language Processing and Speech Recognition, which are based on sequential data processing, significantly benefit from Recurrent Neural Networks (RNNs) 
Recurrent Neural Networks (RNNs) are widely used in applications that require sequential data processing such as natural language processing and speech recognition \cite{Graves2013SpeechRW,enc_dec}.
%Recurrent Neural Networks (RNNs) are now used for many applications such as Natural Language Processing and Speech Recognition, which are based on sequential data processing \cite{}. 
In particular, RNNs equipped with a long-term memory such as Long Short-Term Memory (LSTM) \cite{hochreiter1997long} have proven highly effective and achieved state-of-the-art performance in many tasks \cite{wu_googles_2016,oord_wavenet}. Nevertheless, those RNNs are not without limitation; since they implement their memory using a fixed-size vector, the capacity of their memory is severely restricted and it is hard to have a compartmentalized memory to accurately remember facts about the past \cite{weston_memory_2014}.

To address the limitation of RNNs, researchers have proposed models called Memory-Augmented Neural Networks (MANNs). MANNs are a class of networks equipped with an external memory \cite{santoro_meta-learning_2016}, and they are capable of using individual facts from the past selectively. While MANNs have shown promising results in some (relatively small-scale) experiments, they are not yet practical enough to be widely used in many real-world applications. 

\begin{figure}[t]
  \begin{center}
    \includegraphics[width=\linewidth]{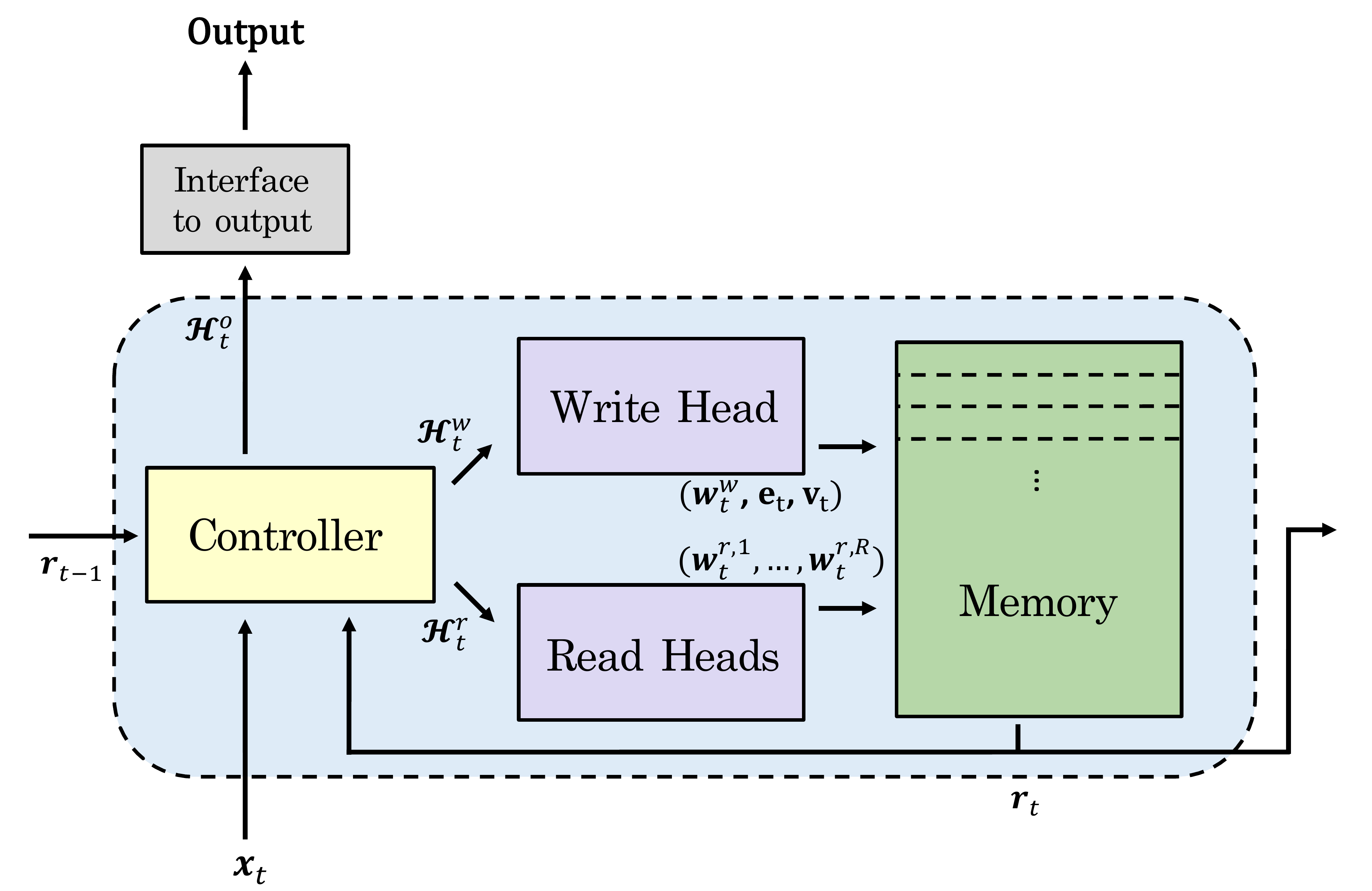} 
    \caption{The architecture of a NTM-based MANN. Note that $\bm{r}_t$ is generated after the read and write operation, and used for $\bm{\mathcal{H}}_t^o$ at $t$.}
    \label{mann_architecture}
  \end{center}
\end{figure}

%In this paper, we present a novel approach to improve the performance of MANNs. 
While there are various types of MANNs, we focus on the MANNs that are based on the Neural Turing Machine (NTM) \cite{graves_neural_2014}. As shown in \figurename\ref{mann_architecture}, a NTM-based MANN consists of a memory and three types of modules implemented using neural networks (NNs), namely, a {\it controller}, {\it read heads}, and a {\it write head}. Among these modules, we focus on the controller, which is the core module that controls how a MANN operates. In most of the previous work on the NTM, the controller is implemented using an LSTM-based RNN because it enables the controller to operate using contextual information. However, it has recently been pointed out that using RNNs for the controller can have negative effects in training the whole model \cite{gulcehre_memory_2017,dntm}. This is mainly because the RNN-based controller has its own memory, and it allows the model to partially solve the given task without using the large external memory, thereby resulting in a suboptimally trained model. 

To address the abovementioned problem, we present a novel type of RNN-based controller that can avoid suboptimal solutions while keeping the ability of using contextual information. Experiments on the Toy tasks \cite{graves_neural_2014,grefenstette_learning_2015,yang_lie-access_2016} and the bAbI task \cite{weston2015towards} demonstrate that our approach substantially improves the performance of MANNs. The main contributions of our work are as follows:

\begin{itemize}
%    \item Organizing a general framework of NTM-based MANNs. 
    \item Introducing a novel type of RNN-based controller for MANNs. This controller utilizes contextual information for controlling the other modules while avoiding its direct use for the outputs of the model.
    \item Demonstrating the effectiveness of the proposed controllers by the experiments on the Toy and bAbI tasks. The experimental results show that the proposed controllers significantly outperform conventional controllers in both tasks.
\end{itemize}
 
%This paper is organized as follows. In section \ref{background_knowledge}, we give a brief introduction to MANNs. We then describe our proposed model in section \ref{proposal}. We present some experimental results in section \ref{experiment}. In section \ref{related_work}, we discuss related work to clarify our contributions and finally, we conclude this paper in section \ref{conclusion}.

\section{Memory-Augmented Neural Networks}
\label{background_knowledge}
\subsection{Model Outline}
\label{NTM_based_MANN}
MANNs are a class of neural networks equipped with an external memory, which is implemented as a set of vectors. Each of the vectors is associated with an address of its memory, and each operation of reading from or writing to the memory is performed with respect to each address. This design of memory use enables a MANN to use a large memory and deal with facts from the past selectively. In this paper, we focus on the NTM-based MANNs, and use the term MANNs to refer to the NTM-based MANNs in what follows. 

%The addresses are called slots, and they 
As shown in \figurename\ref{mann_architecture}, a MANN consists of a memory and three kinds of modules implemented using NNs, namely, a controller, read heads, and a write head. At each time step $t$, these modules follow the procedures from (i) to (iv) as below.
 
\vspace{1.5mm}
\noindent
(i) According to the input to the model $\bm{x}_t \in \mathbb{R}^I$ and the information read from the memory $\bm{r}_{t-1} \in \mathbb{R}^W$, the controller generates two vectors $\bm{\mathcal{H}}_t^r$ and $\bm{\mathcal{H}}_t^w$ to control the read heads and the write head.
%\footnote[1]{In general, $\bm{\mathcal{H}}_t^r$, $\bm{\mathcal{H}}_t^w$, and $\bm{\mathcal{H}}_t^o$ are tuples, e.g. $\bm{\mathcal{H}}_t^o = (\bm{h}_t, \bm{r}_{t})$ in \cite{dnc}, where $\bm{h}_t$ is the output of a NN used in the controller, and $\bm{r}_{t}$ is the vector read from the memory at a time step $t$. However, we treat them as vectors here because we focus on the situation where the number of elements for each tuple is one.}
$I$ is the size of the input vectors, and $W$ is size of each vector in the memory. $\bm{r}_{t-1}$ is defined as $\bm{r}_{t-1} = [\bm{r}_{t-1}^1 ; \, ...; \, \bm{r}_{t-1}^R]$, where $\bm{r}_{t-1}^{i}$ is a vector read from the memory by the $i$th of the $R$ read heads at $t-1$, and the semicolons mean the concatenation of vectors.

\vspace{1.5mm}
\noindent
(ii) According to $\bm{\mathcal{H}}_t^w$ and the memory at $t-1$, $\bm{M}_{t-1} \in \mathbb{R}^{N \times W}$, the write head updates $\bm{M}_{t-1}$ to $\bm{M}_{t}$, where $N$ is the number of addresses of the memory. The write operation is performed as follows:
\begin{equation*}
\label{write_op}
        \bm{M}_t = \bm{M}_{t-1} \odot (\bm{E} - \bm{w}_t^w \bm{e}_t^T) + \bm{w}_t^w \bm{v}_t^T,
\end{equation*}
where all the elements of $\bm{E} \in \mathbb{R}^{N \times W}$ are $1$, and the vectors $\bm{e}_t \in [0, 1]^W$ and $\bm{v}_t \in \mathbb{R}^W$ are used for erasing or adding information in the memory at $t$. $\bm{w}_t^w \in [0, 1]^N$ is a vector which represents the weights for erasing and adding information at each address, where $\sum_j \bm{w}_t^w(j) \leq 1$. $\bm{e}_t$ and $\bm{v}_t$ are generated by a one-layer NN which uses $\bm{\mathcal{H}}_t^w$ as its input.

\vspace{1.5mm}
\noindent
(iii) According to $\bm{\mathcal{H}}_t^r$ and $\bm{M}_{t}$, the read heads read information from the memory, and generate $\bm{r}_t$. %The operation of reading information from the memory is performed as follows:
The read operation of the $i$th read head is performed as follows:
\begin{equation*}
\label{read_op}
        \bm{r}_t^i = \bm{M}_t^T \bm{w}_t^{r, i},
\end{equation*}
where $\bm{w}_t^{r, i} \in [0, 1]^N$ is the weights of the $i$th read head for reading information from each address where $\sum_j \bm{w}_t^{r, i}(j) \leq 1$. After its generation, $\bm{r}_t$ is sent to the controller, and the controller saves it.

\vspace{1.5mm}
\noindent
(iv) According to $\bm{x}_t$, $\bm{r}_{t-1}$, and $\bm{r}_{t}$, the controller generates $\bm{\mathcal{H}}_t^o$, which is the information used for the output of the model.%\footnotemark[1]

\vspace{1.5mm}
\noindent
%%%%%%%%%%Note that the procedures usually follow the order from (i) to (iv), but the order could be different depending on the kinds of models or their implementations. For example, $\bm{r}_t$ in (iv) is not used in some models \cite{Rae2016ScalingMN,dntm}. In this case, the models can perform (i) and (iv) in parallel, leading to efficient implementation.
%There are two possible variations for the above procedures. First, the procedures usually follow the order from (i) to (iv), but the order could be different depending on the kinds of models or their implementations. For example, assume that we do not use $\bm{r}_t$ in the procedure (iv). This is the case in \cite{Rae2016ScalingMN,dntm}. In this case, the models can process (i) and (iv) in parallel, leading to efficient implementation. Second, each module can use the past information on itself. For example, the controller can generate $\bm{\mathcal{H}}_t^w$ according to $\bm{\mathcal{H}}_{1}^w, \, ..., \, \bm{\mathcal{H}}_{t-1}^w$, and the write head can generate $\bm{w}_t^w$ according to $\bm{w}_{1}^w, \, ..., \, \bm{w}_{t-1}^w$.

In the procedures from (i) to (iv), how to generate $\bm{\mathcal{H}}_t^r$, $\bm{\mathcal{H}}_t^w$, $\bm{\mathcal{H}}_t^o$, $\bm{w}_t^w$, and $\bm{w}_t^{r, 1}, \, ..., \, \bm{w}_t^{r, R}$ depends on models and their implementations. In this paper, we use the NTM and the Differentiable Neural Computer (DNC) \cite{dnc} for the models. First, we explain how to generate $\bm{\mathcal{H}}_t^r$, $\bm{\mathcal{H}}_t^w$, and $\bm{\mathcal{H}}_t^o$ for the two models in the next section. We then explain the mechanisms to generate $\bm{w}_t^w$ and $\bm{w}_t^{r, 1}, \, ..., \, \bm{w}_t^{r, R}$ for each model in the following sections.

\subsection{Controller}
\label{controller}
How to generate $\bm{\mathcal{H}}_t^r$, $\bm{\mathcal{H}}_t^w$, and $\bm{\mathcal{H}}_t^o$ is determined by the controller. In this paper, we assume that the baseline controllers for the NTM and the DNC are implemented as $\bm{\mathcal{H}}_t^r = \bm{\mathcal{H}}_t^w = \bm{h}_t$ and $\bm{\mathcal{H}}_t^o = [\bm{h}_t \,;\, \bm{r}_{t}]$, where $\bm{h}_t$ is a vector generated by a NN in the controller according to $\bm{x}_t$ and $\bm{r}_{t-1}$. The same design is used in the original paper of DNC \cite{dnc}.

We consider three types of NNs for the baseline controller: Feedforward Neural Networks (FNNs), Elman Networks (ENs) \cite{elman1990finding}, and LSTMs. We call each type of the controllers a FNN controller, an EN controller, and a LSTM controller. The FNN controller generates $\bm{h}_t$ as follows:
\begin{equation*}
\label{FNN_controller}
    \bm{h}_t = \varphi(\bm{W}_{xh} \bm{x}_t + \bm{W}_{rh} \bm{r}_{t-1} + \bm{b}),
\end{equation*}
where $\varphi$ is an activation function. Similarly, the EN controller generates $\bm{h}_t$ as follows:
\begin{equation}
\label{EN_controller}
    \bm{h}_t = \varphi(\bm{W}_{xh} \bm{x}_t + \bm{W}_{rh} \bm{r}_{t-1} + \bm{W}_{hh} \bm{h}_{t-1} + \bm{b}).
\end{equation}
The LSTM controller generates $\bm{h}_t$ as follows:
\begin{eqnarray}
        \label{LSTM_controller_begin}
        \bm{z}_t &=& \tanh(\bm{W}_{xz} \bm{x}_t + \bm{W}_{rz} \bm{r}_{t-1} + \bm{W}_{hz} \bm{h}_{t-1} + \bm{b}_z), \\
        \label{LSTM_controller_second}
        \bm{i}_t &=& \sigma(\bm{W}_{xi} \bm{x}_t + \bm{W}_{ri} \bm{r}_{t-1} + \bm{W}_{hi} \bm{h}_{t-1} + \bm{b}_i), \\
        \bm{f}_t &=& \sigma(\bm{W}_{xf} \bm{x}_t + \bm{W}_{rf} \bm{r}_{t-1} + \bm{W}_{hf} \bm{h}_{t-1} + \bm{b}_f), \\
        \bm{o}_t &=& \sigma(\bm{W}_{xo} \bm{x}_t + \bm{W}_{ro} \bm{r}_{t-1} + \bm{W}_{ho} \bm{h}_{t-1} + \bm{b}_o), \\
        \label{LSTM_controller_semi_end}
        \bm{c}_t &=& \bm{f}_t \odot \bm{c}_{t-1} + \bm{i} \odot \bm{z}_t, \\
        \label{LSTM_controller_end}
        \bm{h}_t &=& \bm{o}_t \odot \tanh(\bm{c}_t), 
\end{eqnarray}
where $\sigma$ is an activation function for the gating mechanism. In Equations (\ref{EN_controller}--\ref{LSTM_controller_end}), we set $\bm{h}_0 = \bm{c}_0 = \bm{0}$.

\subsection{Neural Turing Machine}
\label{NTM_weight}
For the NTM, $\bm{w}_t^w$ and all of $\bm{w}_t^{r, 1}, \, ..., \, \bm{w}_t^{r, R}$ are generated by the same mechanism. Here we denote them by $\bm{w}_t$.

First, according to $\bm{\mathcal{H}}_t^r = \bm{\mathcal{H}}_t^w = \bm{h}_t$ generated by the controller, the following operation is conducted:
\begin{equation}
\label{content_based_addressing}
    \begin{split}
        \bm{c}_t &= C(\bm{M}, \bm{k}_t, \beta_t)[i] = \frac{\beta_t \exp(\mathcal{K}(\bm{k}_t, \bm{M}[i]))}{\sum_j \beta_t \exp(\mathcal{K}(\bm{k}_t, \bm{M}[j]))},
    \end{split}
\end{equation}
where $\bm{k}_t$ and $\beta$ are generated from $\bm{h}_t$ using a one-layer NN. Note that we use the expression $\bm{M}$ because the write head uses $\bm{M}_{t-1}$, while the read heads use $\bm{M}_{t}$. $\mathcal{K}(\bm{a}, \bm{b})$ is a function which measures the relatedness between two vectors, $\bm{a}$ and $\bm{b}$, and usually implemented by cosine similarity:
\begin{equation*}
\label{cos_sim}
    \begin{split}
        \mathcal{K}(\bm{a}, \bm{b}) &= \frac{\bm{a} \cdot \bm{b}}{|\bm{a}| |\bm{b}|}.
    \end{split}
\end{equation*}
Next, the NTM generates $\bm{w}_t^g$ as follows:
\begin{equation*}
\label{weight_gate}
    \begin{split}
        \bm{w}_t^g = g_t \bm{c}_t + (1 - g_t) \bm{w}_{t-1},
    \end{split}
\end{equation*}
where $g_t \in [0,1]$, and is generated by a one-layer NN. % $\bm{w}_t^g$ represents the template of $\bm{w}_t$.
After that, the following circular convolution is applied to $\bm{w}_t^g$.
\begin{equation*}
\label{conv_shift}
    \begin{split}
        \tilde{\bm{w}_t}[i] &= \sum_{j=0}^{N-1} \bm{w}_t^g[j] \bm{s}_t[i-j],
    \end{split}
\end{equation*}
where $\bm{s}_t \in [0,1]$ represents the amount of shift, and satisfies the condition $\sum_j \bm{s}_t[j] = 1$. $\bm{s}_t$ is generated by a one-layer NN. 
%For example, assume that a vector $(0, 0, 1, 0)$ is shifted. A vector shifted by one is $(0, 0, 0, 1)$, shifted by $2$ is $(1, 0, 0, 0)$, and shifted by $-1$ is $(0, 1, 0, 0)$. In addition, the $\bm{s}_t$ which shifts the vector by $1$ is $(0, 1, 0, 0)$. Note that the upper bound of the shift operation is a hyper parameter. 
Finally, $\bm{w}_t$ is generated as follows: 
\begin{equation*}
\label{weight_fini_ntm}
    \begin{split}
        \bm{w}_t[i] &= \frac{\tilde{\bm{w}_t}[i]^{\gamma_t}}{\sum_j \tilde{\bm{w}_t}[j]^{\gamma_t}},
    \end{split}
\end{equation*}
where $\gamma_t$ satisfies the condition $\gamma_t \geq 1$, and is generated by a one-layer NN. $\gamma_t$ sharpens the element of $\bm{w}_t$.

\subsection{Differentiable Neural Computer}
\label{DNC}
In the DNC, $\bm{w}_t^w$ and $\bm{w}_t^{r, 1}, \, ..., \, \bm{w}_t^{r, R}$ are generated by different mechanisms.

First, we explain the write operation. The following operation is conducted.
\begin{equation*}
\label{dnc_write_1}
    \bm{\psi}_t = \prod_{i=1}^R (\bm{1} - f_t^i\bm{w}_{t-1}^{r, i}),
\end{equation*}
where $f_t^i \in [0,1]$ is a scalar generated by a one-layer NN for each read head. $\bm{\psi}_t \in [0,1]^N$ represents how much each address will not be freed by the free gates, $f_t^i$. According to $\bm{\psi}_t$, the usage vector is defined as follows:
\begin{equation*}
\label{dnc_write_2}
\bm{u}_t = (\bm{u}_{t-1} + \bm{w}_{t-1}^w - \bm{u}_{t-1} \odot \bm{w}_{t-1}^w) \odot \bm{\psi}_t,
\end{equation*}
where each element of $\bm{u}_t$ indicates the degree to which the address is used, and the nearer it is to $1$, the higher the degree is. %Intuitively, locations are used if they have been retained by the free gates ($\bm{\psi}_t[i] \approx 1$), and were either already in use or have just been written to. 
After that, $\bm{\phi}_t \in \mathbb{Z}^N$ is defined. Each element of $\bm{\phi}_t$ represents an index, and they are sorted by ascending order of usage. By using $\bm{\phi}_t$, the allocation weighting, which is used to provide new addresses for writing is generated as follows:
\begin{equation*}
\label{dnc_write_3}
\bm{a}_t[\bm{\phi}_t[j]] = (1 - \bm{u}_t[\bm{\phi}_t[j]]) \prod_{i=1}^{j-1} \bm{u}_t[\bm{\phi}_t[i]].
\end{equation*}
According to $\bm{a}_t$, the actual address used for the write operation is defined as follows:
\begin{equation*}
\label{dnc_write_4}
\bm{w}_t^w = g_t^w[g_t^a\bm{a}_t+(1-g_t^a)\bm{c}_t^w],
\end{equation*}
where $g_t^w \in [0,1]$ and $g_t^a \in [0,1]$ are scalars generated by a one-layer NN. $\bm{c}_t^w$ is a vector generated as $\bm{c}_t$ in \equationname(\ref{content_based_addressing}).

The read operation is conducted using a temporal link matrix, $\bm{L}_t \in [0,1]^{N \times N}$. This matrix holds the order of written addresses. In the DNC, the following operation is conducted according to $\bm{w}_t^w$:
\begin{equation*}
\label{dnc_write_5}
\bm{p}_t = (1 - \sum_i \bm{w}_t^w[i]) \bm{p}_{t-1} + \bm{w}_t^w,
\end{equation*}
where $\bm{p}_0 = \bm{0}$. $\bm{p}_t$ basically represents the addresses where the write operation is conducted at $t$, while it holds the recently written addresses when the write operation is not conducted. $\bm{L}_t$ tracks the write operation by the following operation:
\begin{equation*}
\label{dnc_write_6}
\bm{L}_t[i,j] = (1 - \bm{w}_t^w[i] - \bm{w}_t^w[j]) \bm{L}_{t-1}[i,j] + \bm{w}_t^w[i]\bm{p}_{t-1}[j],
\end{equation*}
where $\bm{L}_0[i,j] = 0, \forall i,j$ and $\bm{L}_t[i,i] = 0$. By using this matrix, vectors $\bm{f}^i_t$ and $\bm{b}^i_t$ are defined as follows:
\begin{eqnarray*}
    \bm{f}_t^i &=& \bm{L}_t \cdot \bm{w}_t^{r,i}, \\
    \bm{b}_t^i &=& \bm{L}_t^T \cdot \bm{w}_t^{r,i}.
\end{eqnarray*}
The two vectors represent the addresses where the write operations are conducted before and after the location $\bm{w}_t^{r,i}$ is written. Finally, the addresses for the read operation is defined as follows:
\begin{equation*}
\label{dnc_write_10}
\bm{w}_t^{r,i} = \bm{\pi}_t^i[1]\bm{b}_t^i + \bm{\pi}_t^i[2]\bm{c}_t^{r,i} + \bm{\pi}_t^i[3]\bm{f}_t^i,
\end{equation*}
where $\bm{\pi}_t^{r,i} \in [0,1]^3$ is generated according to $\bm{h}_t$ using a one-layer NN which uses a softmax function for its activation function. We do not use the sparse link matrix for the DNC in this paper.

\section{Partially Non-Recurrent Controllers}
\label{proposal}
RNN-based controllers enable MANNs to utilize contextual information for controlling the other modules. This is usually beneficial for the models, and most of the studies on MANNs adopt a RNN-based controller for their models. However, the use of the memory in the RNN-based controller potentially has a negative effect for the training of the models because the output of the controller $\bm{h}_t$ is used for $\bm{\mathcal{H}}_t^o$, which allows the model to directly solve the given tasks using the (small) memory in the controller.

\begin{figure}[t]
  \begin{center}
    \includegraphics[width=\linewidth]{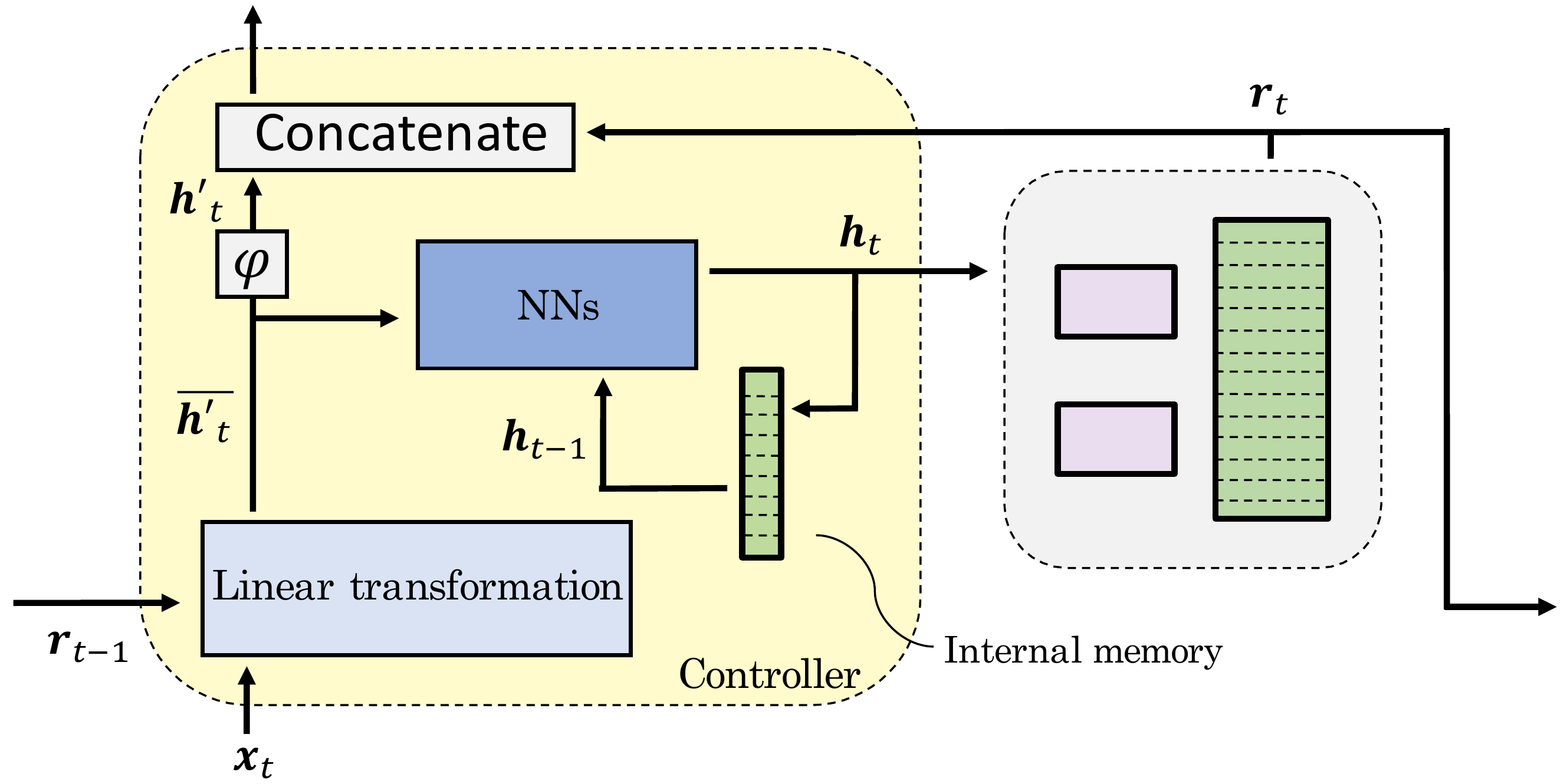}
    \caption{The architecture of the partially non-recurrent controller.}
    \label{proposed_controller_fig}
  \end{center}
\end{figure}

In this paper, we propose a novel type of RNN-based controller that is partially non-recurrent and avoids the direct use of its internal memory for solving the task, while keeping the ability of using contextual information in controlling the other modules. As shown in \figurename\ref{proposed_controller_fig}, the outputs of the proposed RNN-based controller are $\bm{\mathcal{H}}_t^r = \bm{\mathcal{H}}_t^w = \bm{h}_t$ and $\bm{\mathcal{H}}_t^o = [\bm{h}'_t \,;\, \bm{r}_{t}]$, where $\bm{h}_t$ is the vector generated in the same way as usual RNN-based controllers, and $\bm{h}'_t$ is the vector generated without using the memory in the controllers. For the EN controller, $\bm{h}'_t$ is generated as follows:
\begin{eqnarray}
        \label{proposed_EN_controller_begin}
        \bar{\bm{h}'_t} &=& \bm{W}_{xh} \bm{x}_t + \bm{W}_{rh} \bm{r}_{t-1} + \bm{b}, \\
        \bm{h}'_t &=& \varphi(\bar{\bm{h}'_t}).
        \label{proposed_EN_controller_end}
\end{eqnarray}
Then, $\bm{h}_t$ is generated by using $\bar{\bm{h}'_t}$ as follows:
\begin{equation}
\label{proposed_EN_controller_for_control}
        \bm{h}_t = \varphi(\bm{W}_{hh} \bm{h}_{t-1} + \bar{\bm{h}'_t}).
\end{equation}
Note that the number of parameters used in \equationname(\ref{proposed_EN_controller_begin}) and \equationname(\ref{proposed_EN_controller_for_control}) is same as that used in \equationname(\ref{EN_controller}). %In \equationname(\ref{proposed_EN_controller_for_control}), we set $\bm{h}_0 = \bm{0}$.

%\begin{equation}
%\label{proposed_EN_controller_begin}
%    \bm{h}'_t = \varphi(\bm{W}_{xh} \bm{x}_t + \bm{W}_{rh} \bm{r}_{t-1} + \bm{b}).
%\end{equation}

%Note that the parameters denoted by the same vectors and matrices in \equationname(\ref{EN_controller}) and \equationname(\ref{proposed_EN_controller_begin}) are shared. This sharing enables our proposed RNN-based controller to generate $\bm{h}'_t$ without increasing the number of parameters. 

Similarly, for the LSTM controller, $\bm{h}'_t$ is generated in the same manner as \equationname(\ref{proposed_EN_controller_end}) by using the following $\bar{\bm{h}'_t}$:
%Similarly, for the LSTM controller, $\bm{h}'_t$ is computed as follows:
%\vspace{-4mm}
\begin{equation}
\label{proposed_LSTM_controller_begin}
        \bar{\bm{h}'_t} = \bm{W}_{xz} \bm{x}_t + \bm{W}_{rz} \bm{r}_{t-1} + \bm{b}_{z}.
\end{equation}
Then, $\bm{h}_t$ is generated according to Equations (\ref{LSTM_controller_second}--\ref{LSTM_controller_end}) by using the following $\bm{z}_t$:
\begin{equation}
\label{proposed_LSTM_controller_z}
        \bm{z}_t = \tanh(\bm{W}_{hz} \bm{h}_{t-1} + \bar{\bm{h}'_t}).
\end{equation}
Again, the number of parameters used in \equationname(\ref{proposed_LSTM_controller_begin}) and \equationname(\ref{proposed_LSTM_controller_z}) is the same as that used in \equationname(\ref{LSTM_controller_begin}). %In \equationname(\ref{proposed_EN_controller_for_control}), we set $\bm{h}_0 = \bm{0}$.
%Again, the parameters denoted by the same vectors and matrices in \equationname(\ref{proposed_LSTM_controller_begin}) and \equationname(\ref{LSTM_controller_begin}--\ref{LSTM_controller_end}) are shared.

%A key point to make use of the proposed RNN-based controllers is the implementation of $\bm{\mathcal{H}}_t^o$. We empirically find that the models with the proposed controllers do not perform well in case we remove $\bm{r}_{t}$ from $\bm{\mathcal{H}}_t^o$, while this affects less for the models with the usual RNN-based controllers. This is because the proposed controller cannot use the past information depending on $\bm{x}_t$ at $t$ in this case, while the usual RNN-based controllers can use the internal memory of the controller.

Although in this paper we apply our proposal only to the EN and the LSTM controller, it can be applied to other types of RNN-based controllers such as the RNN-based controllers based on gated recurrent units \cite{cho_learning_2014}.

\section{Experiments}
\label{experiment}
\subsection{General Settings}
To evaluate the performance of our proposed RNN-based controller, we carry out experiments on two sets of tasks, the Toy and bAbI tasks. On both tasks, we compare the performance of the NTM and the DNC with the FNN controller, the EN controller, the proposed EN controller, the LSTM controller, and the proposed LSTM controller. The network settings of the NTM and the DNC are described in the following sections separately because they are different on the two tasks, except for the upper bound of the shift operation of the NTM, which is $\pm3$. We also carry out experiments on a one-layer EN and LSTM with $128$ hidden units to evaluate how well the RNNs in RNN-based controllers can solve the tasks. For parameter optimization, we use RMSProp \cite{graves2013generating} with a learning rate of 0.0001 and a momentum of 0.9. The training is performed in an online manner, and during backpropagation we clip all gradient values by the global norm with a threshold of 5. 
%to the range [-10, 10]. %by the global norm with threshold 5. 
When evaluating the models, we use the best model parameters in terms of validation loss, and we report average scores of ten individual models with different parameter initializations for each experimental setting.
% It took about sixteen hours to 

\subsection{Toy Tasks}
\label{toy_tasks}
\textbf{Settings.} Following the previous work \cite{graves_neural_2014,grefenstette_learning_2015,yang_lie-access_2016}, we use six kinds of the Toy tasks described in \tablename\ref{tasks_detail}. On each task, the models receive a sequence of nine-dimensional binary vectors, and they are required to output an appropriate sequence of vectors. The ninth element of each vector indicates the end of the input and output sequences. 
%, and only on \textsc{Repeat Copy}, the models are also required to output this flag after each repetition. 
The number of the input vectors is indicated by $T$, and it is chosen randomly for each input sequence. We adopt $T \in [1, 20]$ for all tasks except for \textsc{Repeat Copy}, for which we adopt $T \in [1, 10]$ and $M \in [1, 10]$, where $M$ is the randomly chosen number of repetitions. We use ten different training datasets, each of which consists of $1,000,000$ sequences for each individual model. The sizes of test and validation data are $10,000$ and $1,000$, respectively, and we validate the models for every $1,000$ training iterations. We use the unit size of $128$ for all of the controllers. The memory size of the NTM and the DNC is $128 \times 20$, and the number of the read heads is set to one for all of the tasks except for \textsc{Priority Sort}, for which we use the models with four read heads. For evaluation, we use the average bit error rates of the output of the ten models.

\begin{table}[t]
%\begin{table*}[htb]
    \begin{center}
        \scalebox{0.78}[0.78]{
  \begin{tabular}{l|lll} \hline
      & Input & Output \\ \hline \hline
      \textsc{Copy} & $\bm{a}_1 \bm{a}_2 \bm{a}_3 \bm{a}_4 \, ... \, \bm{a}_T$ & $\bm{a}_1 \bm{a}_2 \bm{a}_3 \bm{a}_4 \, ... \, \bm{a}_T$ \\
      \textsc{Reverse} & $\bm{a}_1 \bm{a}_2 \bm{a}_3 \bm{a}_4 \, ... \, \bm{a}_T$ & $\bm{a}_T \bm{a}_{T-1} \bm{a}_{T-2} \bm{a}_{T-3} \, ... \, \bm{a}_1$ \\
      \textsc{Bigram Flip} & $\bm{a}_1 \bm{a}_2 \bm{a}_3 \bm{a}_4 \, ... \, \bm{a}_T$ & $\bm{a}_2 \bm{a}_1 \bm{a}_4 \bm{a}_3 \, ... \, \bm{a}_T$ \\
      \textsc{Odd First} & $\bm{a}_1 \bm{a}_2 \bm{a}_3 \bm{a}_4 \, ... \, \bm{a}_T$ & $\bm{a}_1 \bm{a}_3 \, ... \, \bm{a}_2 \bm{a}_4 \, ... $ \\
      \textsc{Repeat Copy} & $\bm{a}_1 \bm{a}_2 \bm{a}_3 \bm{a}_4 \, ... \, \bm{a}_T M$ & $\bm{a}_1 \, ... \, \bm{a}_T \, ... \, \bm{a}_1 \, ... \, \bm{a}_T \, (M \, times)$ \\
      \textsc{Priority Sort} & $\bm{a}_1^5 \bm{a}_2^{10} \bm{a}_3^1 \bm{a}_4^T \, ... \, \bm{a}_T^4$ & $\bm{a}_3^1 \bm{a}_7^{2} \bm{a}_8^3 \bm{a}_T^4 \, ... \, \bm{a}_4^T$ \\ \hline
  \end{tabular}
  }
  \caption{Details of the Toy tasks. The superscripts in \textsc{Priority Sort} is the priorities defined on the dataset, and implicitly attached to each kind of vectors.}
  \label{tasks_detail}
  \end{center}
\end{table}

\begin{figure}[t]
  \begin{center}
        \subfigure[NTM with the LSTM controller]{\includegraphics[scale=0.32]{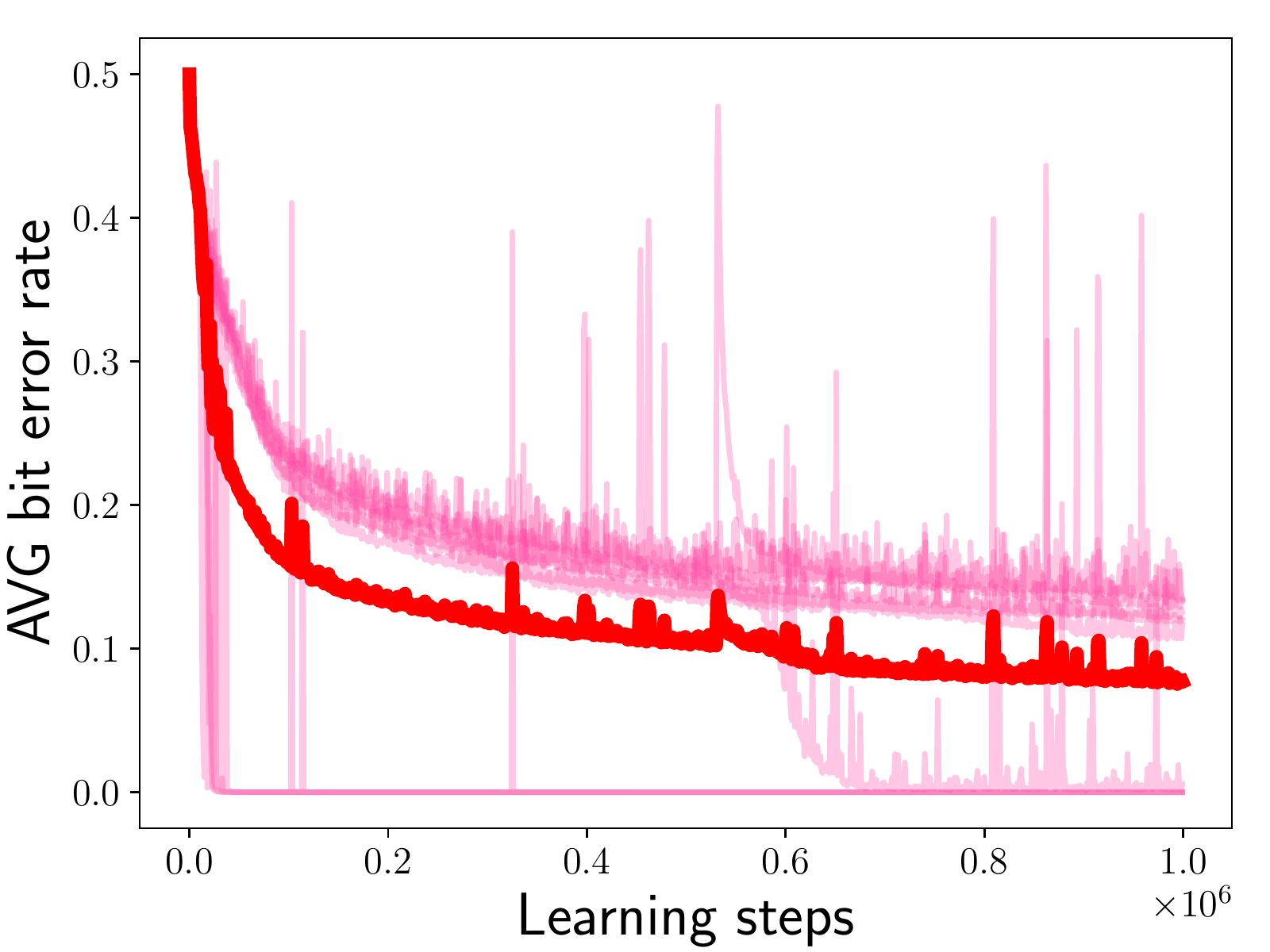} \label{detailed_learning_curve_1}}
        \subfigure[NTM with the proposed LSTM controller]{\includegraphics[scale=0.32]{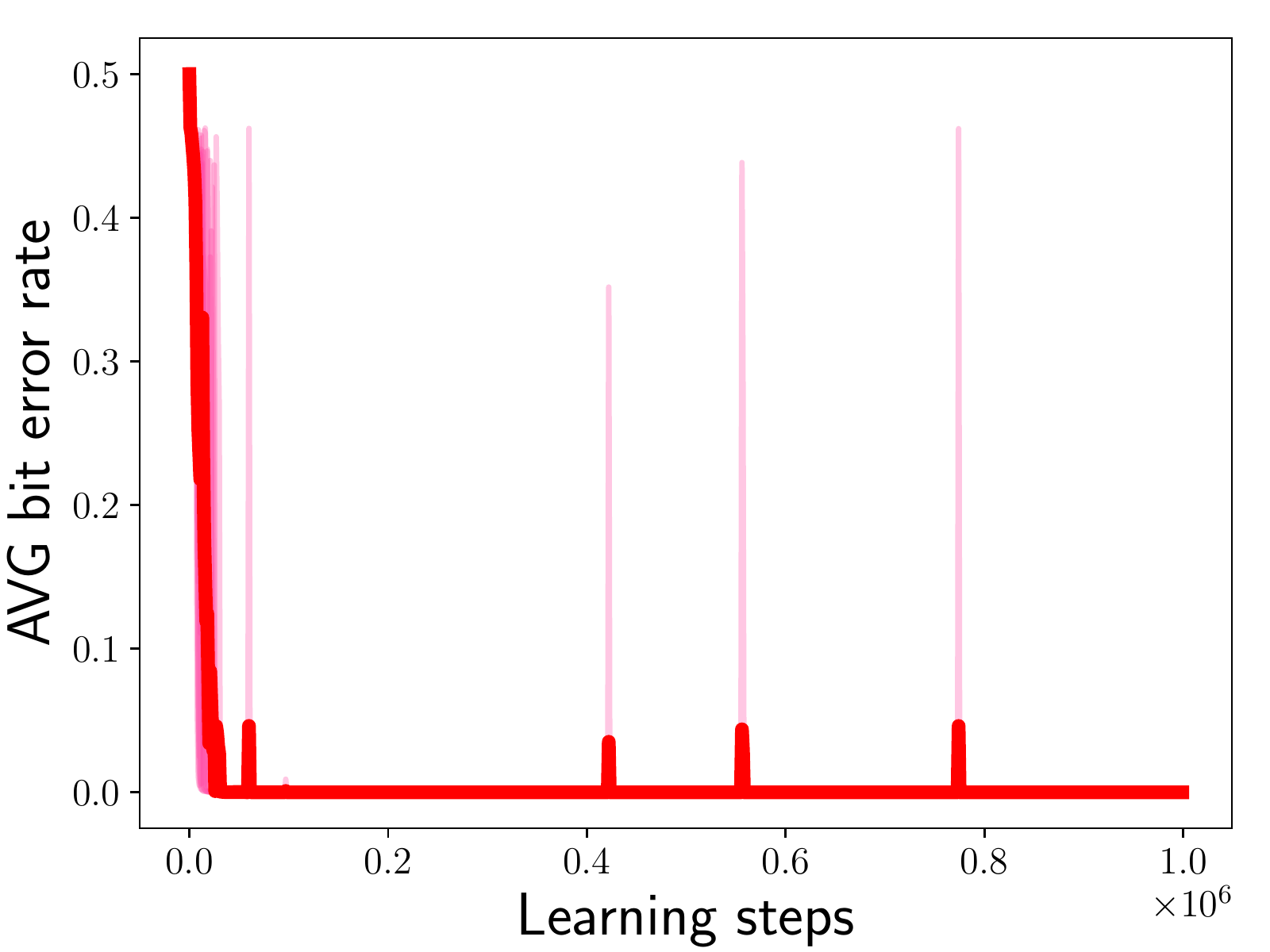} \label{detailed_learning_curve_2}}
    \caption{Learning curves of the NTM with the LSTM and the proposed LSTM controller on \textsc{Copy}. Bold lines represent the average learning curves, and each light colored line represents the learning curve of the one of the ten individual experiments.}
    \label{detailed_learning_curves}
  \end{center}
\end{figure}

\begin{figure}[t]
  \begin{center}
        \subfigure[NTM with the LSTM controller]{\includegraphics[scale=0.33]{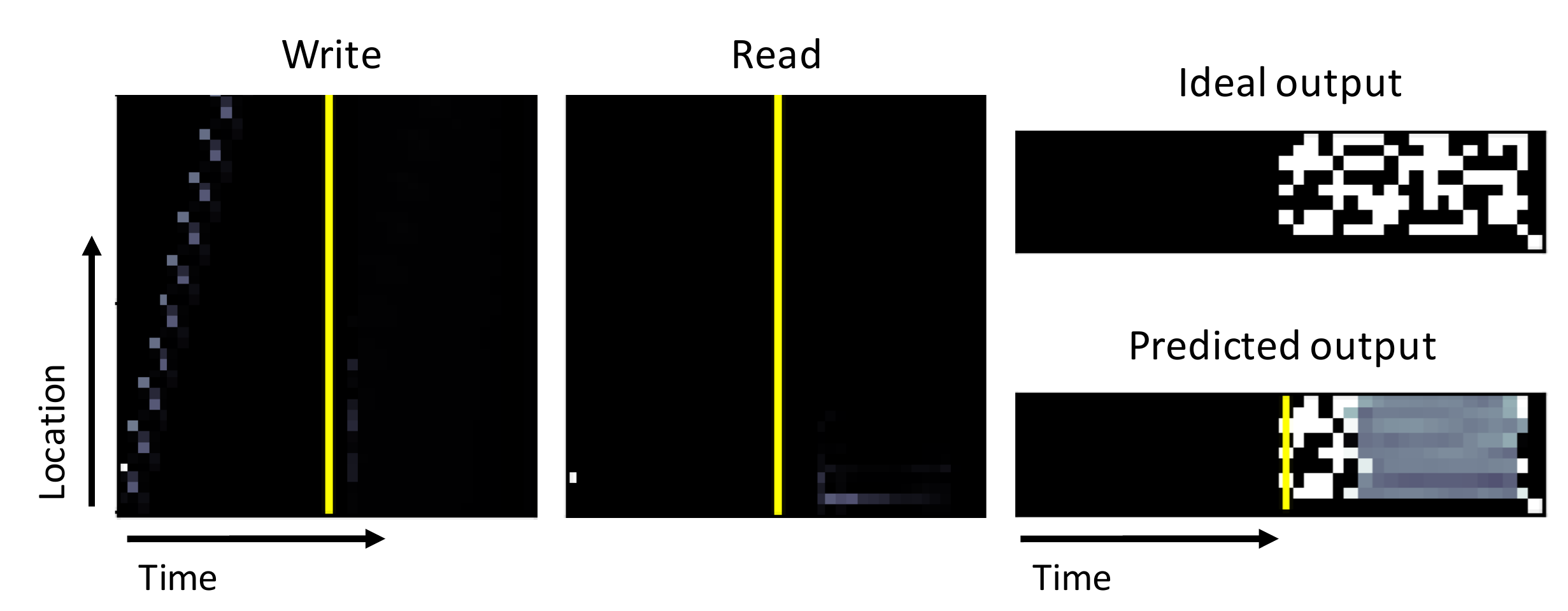} \label{concrete_output1}}
        \subfigure[NTM with the proposed LSTM controller]{\includegraphics[scale=0.33]{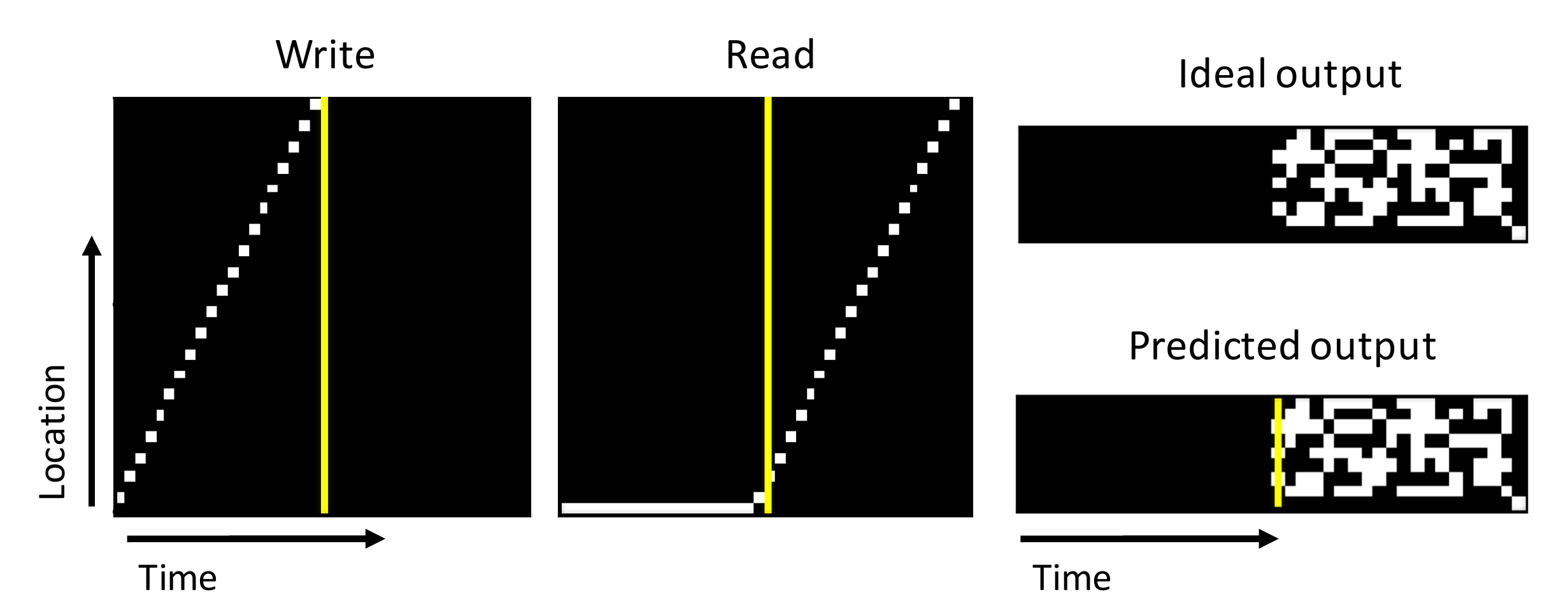} \label{concrete_output2}}
        \caption{Examples of output and memory use of the NTM with the LSTM and the proposed LSTM controller on \textsc{Copy}. In the figure of Read and Write, the white addresses are read or written. The yellow line is the border between the input phase and the output phase. Note that only a subset of memory locations are shown.}
    \label{concrete_output}
  \end{center}
\end{figure}

\vspace{1mm}
\noindent
\textbf{Discussions on the test results.}
\tablename\ref{toy_result} shows the experimental results on the Toy tasks. As shown in \tablename\ref{toy_result}, the NTM or the DNC with one of the proposed controllers achieves the lowest average bit error rates on all of the tasks. \figurename\ref{detailed_learning_curves} illustrates why the models with the proposed controllers achieve the best results. In \figurename\ref{detailed_learning_curve_1}, some of the ten models converge insufficiently, while all of the models converge successfully in \figurename\ref{detailed_learning_curve_2}. Also, we show examples of the output and the memory use of the NTM with the LSTM and the proposed LSTM controller on \textsc{Copy} in \figurename\ref{concrete_output}. As seen in \figurename\ref{concrete_output2}, the NTM with the proposed LSTM controller predicts the perfect output, making an appropriate use of the external memory, while the output of the NTM with the LSTM controller (\figurename\ref{concrete_output1}) is far from perfect. An interesting observation is that the output of the NTM with the LSTM controller is partially correct although it does not read from the address where it wrote the information in the past. This phenomenon occurs because the model solves the task using the small memory in the controller directly as we hypothesized, and the phenomenon is seen for all of the insufficiently converged NTMs with the LSTM controller as shown in \figurename\ref{detailed_learning_curve_1}.

\begin{figure}[t]
  \begin{center}
        \subfigure[Failer case]{\includegraphics[scale=0.33]{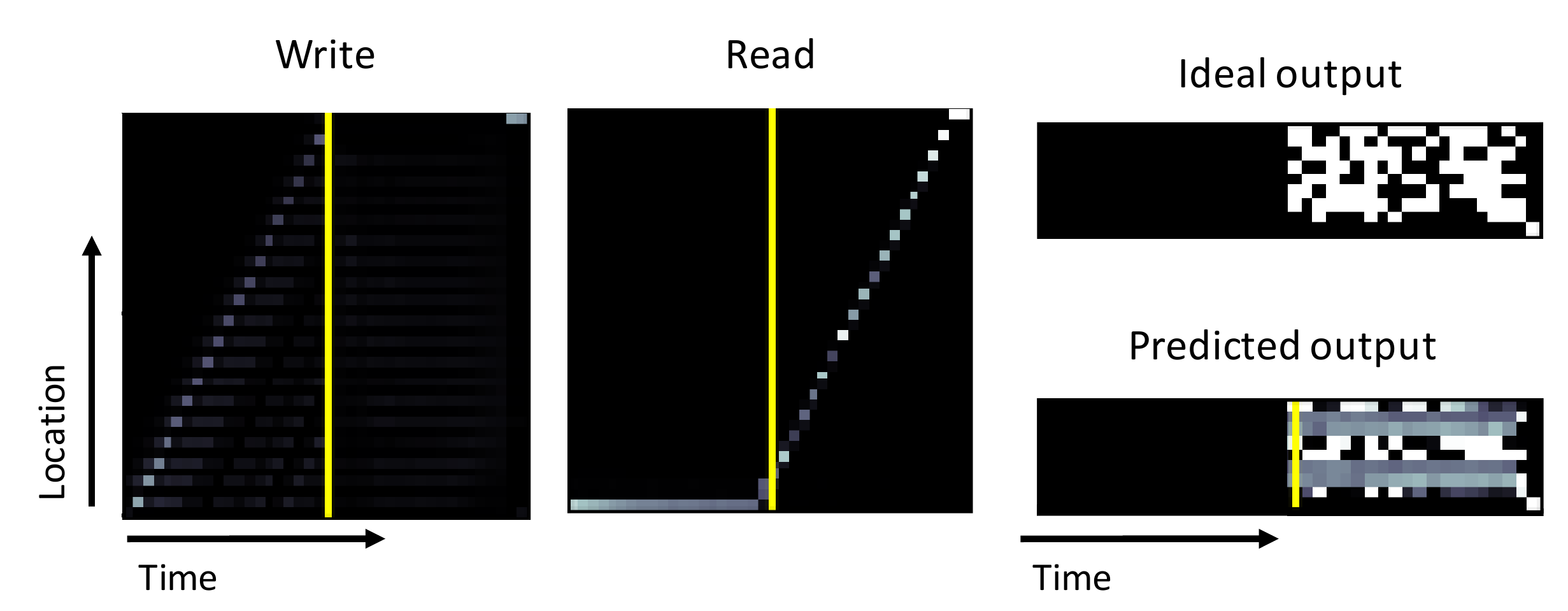} \label{bad_output1}}
        \subfigure[Successful case]{\includegraphics[scale=0.33]{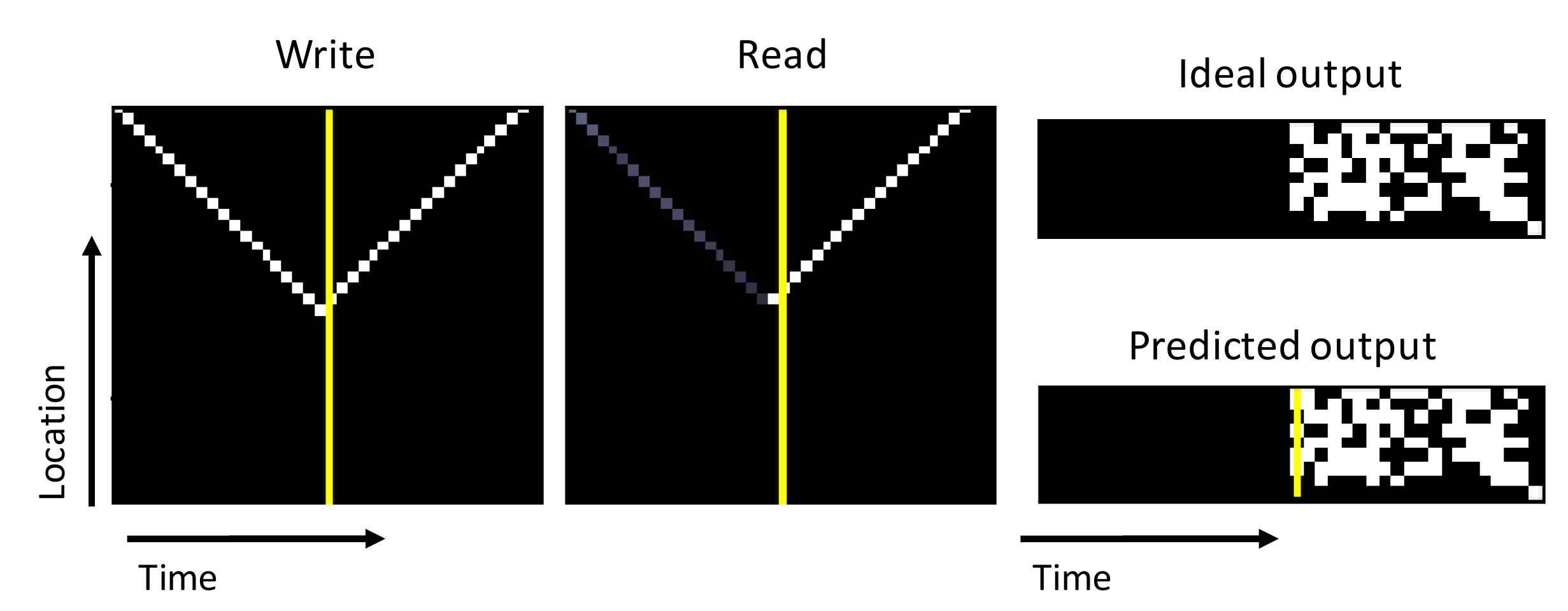} \label{bad_output2}}
    \caption{Failed and successful examples of output and memory use of the NTM with the proposed EN controller on \textsc{Reverse}. In the figure of Read and Write, the white addresses are read or written. The yellow line is the border between the input phase and the output phase, and only a subset of memory locations are shown. Note that in \textsc{Reverse}, successfully trained models read written information reversely as seen in \figurename\ref{bad_output2}}%, while the correct solution on \textsc{Copy} is different as seen in \figurename\ref{concrete_output1}}
    \label{bad_output}
  \end{center}
\end{figure}

Nonetheless, there are situations where the proposed controllers perform worse than the other controllers. Among them, we focus on the results of the NTM with the proposed EN controller on \textsc{Reverse} because both of the average bit error rate and the number of trained models which completely solved the task are worse than those of the NTM with the conventional controller only for this case. We show failed and successful examples of output and memory use of the NTM with the proposed EN controller on \textsc{Reverse}. In the experiment, we find that the NTM with the proposed EN controller tends to converge to the solution shown in \figurename\ref{bad_output1} or similar ones. In \figurename\ref{bad_output1}, the model predicts partially correct outputs although it does not read the written information reversely. %as shown in \figurename\ref{bad_output2}. 
Because the proposed EN controller cannot use the internal memory of the controller directly to solve tasks, the phenomenon that the model even partially solves the task cannot occur without using the external memory. In \figurename\ref{bad_output1}, we can see that the write operation is conducted on multiple addresses at each time step, while the read operation is conducted on just one address. In addition, the read operation in the input phase is conducted only on one specific address. These observations suggest that the model converges to a local optimum where it holds the partial contextual information using the internal memory of the controller, and send it to the output using multiple memory locations. This type of local optimum tends to occur with the proposed controllers but not with the FNN controller and the standard RNN-based controllers. The FNN controller does not suffer from the second phenomenon because they do not have the internal memory of the controller, and the usual RNN-based controllers do not suffer from the two phenomena because they can directly use the internal memory of the controller for the output of the model.

\begin{table*}[htb]
    \begin{center}
        \scalebox{0.7}[0.7]{
  \begin{tabular}{c|c|c|ccccc|ccccc} \hline
      & EN & LSTM  & \multicolumn{5}{|c|}{NTM} & \multicolumn{5}{|c}{DNC} \\ \cline{4-13}
      &  &  & FNN & EN & proposed EN & LSTM & proposed LSTM & FNN & EN & proposed EN & LSTM & proposed LSTM \\ \hline \hline
      \textsc{Copy}   & 37.5 (0) & 21.8 (0) & \textbf{\underline{0.0}} (10) & \textbf{\underline{0.0}} (9) & \textbf{\underline{0.0}} (10) & 11.0 (3) & \textbf{\underline{0.0}} (10) & 2.8 (9) & \textbf{\underline{0.0}} (9) & \textbf{\underline{0.0}} (10) & 13.2 (3) & 3.3 (7) \\
      \textsc{Reverse}   & 25.8 (0) & 13.2 (0) & 10.0 (6) & 2.0 (8) & 15.2 (2) & 8.9 (2) & 7.8 (3) & 2.2 (9) & 7.4 (5) & \textbf{\underline{0.0}} (10) & 7.7 (1) & 13.5 (3) \\
      \textsc{Bigram Flip}  & 37.1 (0) & 23.2 (0) & 4.5 (3) & 2.0 (8) & \textbf{\underline{0.0}} (10) & 9.6 (5) & \textbf{\underline{0.0}} (10) & 2.8 (5) & 0.8 (8) & 1.7 (8) & 9.6 (3) & 7.0 (6) \\
      \textsc{Odd First}  & 36.0 (0) & 13.1 (0) & 3.2 (1) & 2.9 (6) & 0.2 (7) & 5.7 (3) & \textbf{\underline{0.0}} (10) & 18.0 (0) & 4.7 (2) & 2.7 (6) & 10.1 (1) & 10.6 (1) \\
      \textsc{Repeat Copy}  & 15.6 (0) & 7.7 (0) & 1.0 (8) & \textbf{\underline{0.0}} (10) & \textbf{\underline{0.0}} (10) & 0.9 (5) & \textbf{\underline{0.0}} (10) & 1.5 (8) & 0.3 (9) & \textbf{\underline{0.0}} (10) & \textbf{\underline{0.0}} (10) & \textbf{\underline{0.0}} (10) \\
      \textsc{Priority Sort}  & 30.0 (0) & 14.7 (0) & 11.2 (0) & 7.4 (0) & 9.3 (0) & 7.5 (0) & 8.2 (0) & 12.1 (0) & 6.7 (0) & 9.0 (0) & 8.3 (0) & \textbf{\underline{4.4}} (0) \\ \hline
  \end{tabular}
  }
  \caption{Average bit error rates on the Toy tasks. Bold results are the best ones for each task. The bracketed numbers are the number of the individual models of the ten which completely solved (achieved 0.0\% of bit error rate) the tasks.}
  \label{toy_result}
  \end{center}

    \begin{center}
        \scalebox{0.75}[0.75]{
  \begin{tabular}{l|c|c|ccccc|ccccc} \hline
      & EN & LSTM  & \multicolumn{5}{|c|}{NTM} & \multicolumn{5}{|c}{DNC} \\ \cline{4-13}
      &  &  & FNN & EN & proposed EN & LSTM & proposed LSTM & FNN & EN & proposed EN & LSTM & proposed LSTM \\ \hline \hline
      1: 1 supporting fact & 76.9 & 30.0 & 1.1 & 33.7 & 26.5 & 6.1 & 1.4 & 12.7 & 56.8 & 36.7 & 11.7 & \textbf{0.1} \\
      4: 2 argument rels. & 66.8 & 1.4 & 0.7 & 18.3 & 12.9 & 0.5 & \textbf{0.1} & 2.3 & 32.7 & 3.9 & 0.6 & 0.2 \\
      9: simple negation & 80.0 & 17.8 & 7.9 & 14.3 & 22.1 & 5.5 & 4.3 & 13.8 & 34.6 & 16.9 & 10.8 & \textbf{0.7} \\
      10: indefinite knowl. & 83.8 & 31.3 & 13.6 & 22.0 & 32.4 & 20.2 & 11.2 & 25.0 & 43.8 & 25.8 & 22.7 & \textbf{2.7} \\
      11: basic coreference & 71.1 & 10.8 & 0.5 & 19.2 & 16.6 & 1.5 & 0.4 & 8.7 & 39.9 & 24.6 & 2.9 & \textbf{0.1} \\
      14: time reasoning & 89.8 & 55.7 & 33.8 & 44.7 & 54.3 & 41.8 & 26.4 & 48.1 & 61.2 & 51.2 & 58.6 & \textbf{15.4} \\ \hline
      Mean err. & 78.1 & 24.5 & 9.6 & 25.4 & 27.5 & 12.6 & 7.3 & 19.1 & 44.8 & 26.5 & 15.0 & \textbf{3.2} \\ \hline
  \end{tabular}
  }
  \caption{Average error rates on the bAbI task. Bold results are the best ones for each task.}
  \label{mean_babi_result}
  \end{center}
%  \begin{center}
%        \scalebox{0.75}[0.75]{
%  \begin{tabular}{l|c|c|ccccc|ccccc} \hline
%      & EN & LSTM  & \multicolumn{5}{|c|}{NTM} & \multicolumn{5}{|c}{DNC} \\ \cline{4-13}
%      &  &  & FNN & EN & proposed EN & LSTM & proposed LSTM & FNN & EN & proposed EN & LSTM & proposed LSTM \\ \hline \hline
%      1: 1 supporting fact & 47.4 & 25.6 & \textbf{0.0} &  &  & 0.2 & \textbf{0.0} & 2.0 &  &  & \textbf{0.0} & 0.1 \\
%      4: 2 argument rels. & 42.8 & 0.4 & 0.5 &  &  & \textbf{0.0} & \textbf{0.0} & 6.1 &  &  & \textbf{0.0} & \textbf{0.0} \\
%      9: simple negation & 27.7 & 16.2 & 0.5 &  &  & 0.5 & \textbf{0.1} & 5.6 &  &  & \textbf{0.1} & 1.3 \\
%      10: indefinite knowl. & 58.7 & 73.0 & 0.7 &  &  & 0.4 & 0.1 & 9.6 &  &  & \textbf{0.2} & 2.1 \\
%      11: basic coreference & 75.0 & 92.5 & \textbf{0.0} &  &  & \textbf{0.0} & \textbf{0.0} & 2.7 &  &  & \textbf{0.0} & \textbf{0.0} \\
%      14: time reasoning & 19.6 & 46.3 & 10.5 &  &  & 7.4 & \textbf{3.3} & 31.4 &  &  & 4.0 & 6.9 \\ \hline
%      Mean err. & 44.1 & 21.7 & 2.0 &  &  & 1.4 & \textbf{0.6} & 9.6 &  &  & 0.7 & 1.7 \\ \hline
%  \end{tabular}
%  }
%  \caption{Error rates of the models with the best parameter values on bAbI task. We choose the best parameter values in terms of the mean error rates o%f each individual model on the joint task.}
%  \label{best_babi_result}
%  \end{center}
%\end{table*}

%\begin{table*}[htb]
\end{table*}

\begin{figure}[tbp]
  \begin{center}
      \subfigure[NTM w/ FNN controller]{\includegraphics[scale=0.25, bb=0 0 461 346]{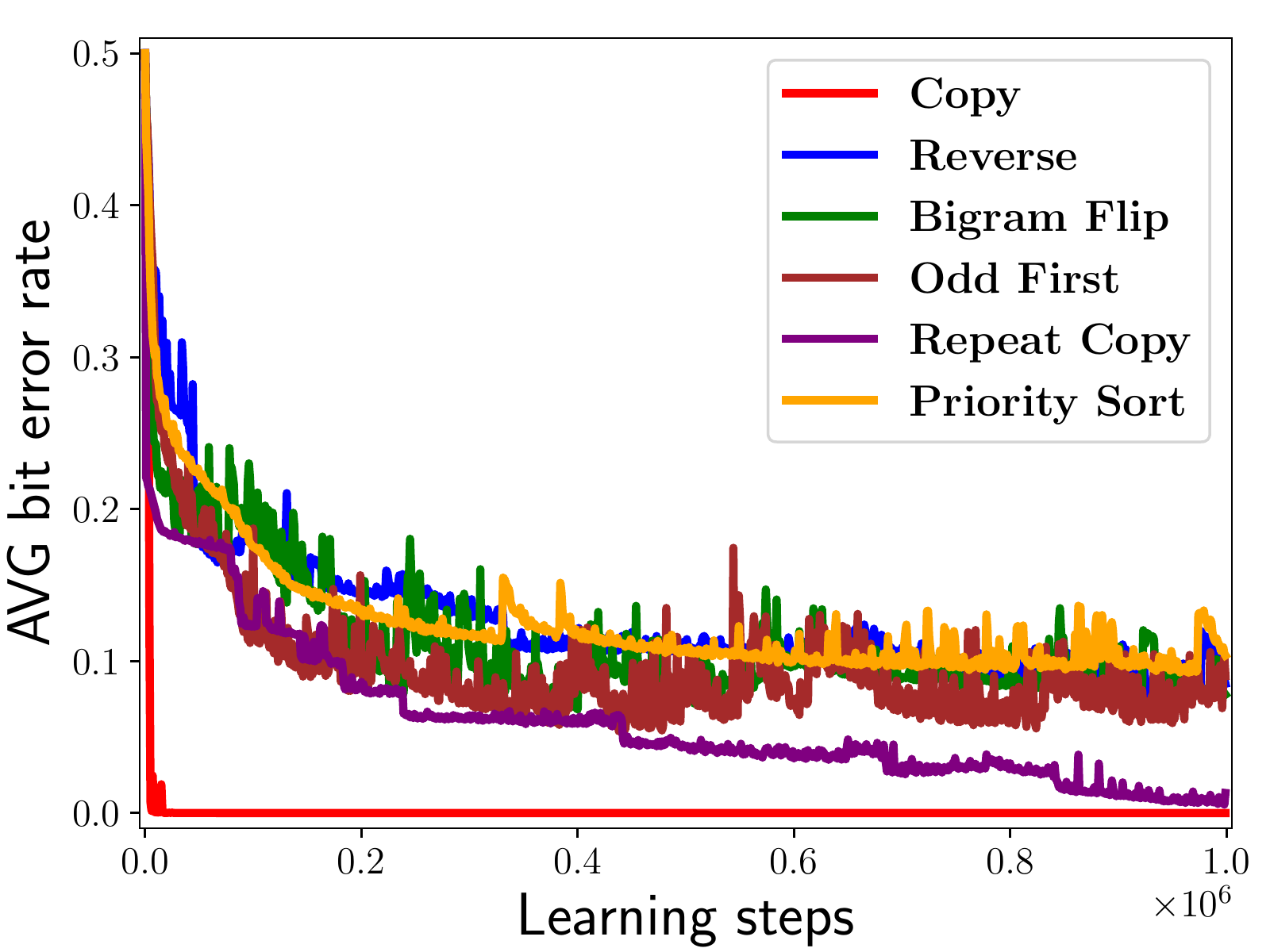} \label{ntm_dense_learning_curve}}
        \subfigure[DNC w/ FNN controller]{\includegraphics[scale=0.25, bb=0 0 461 346]{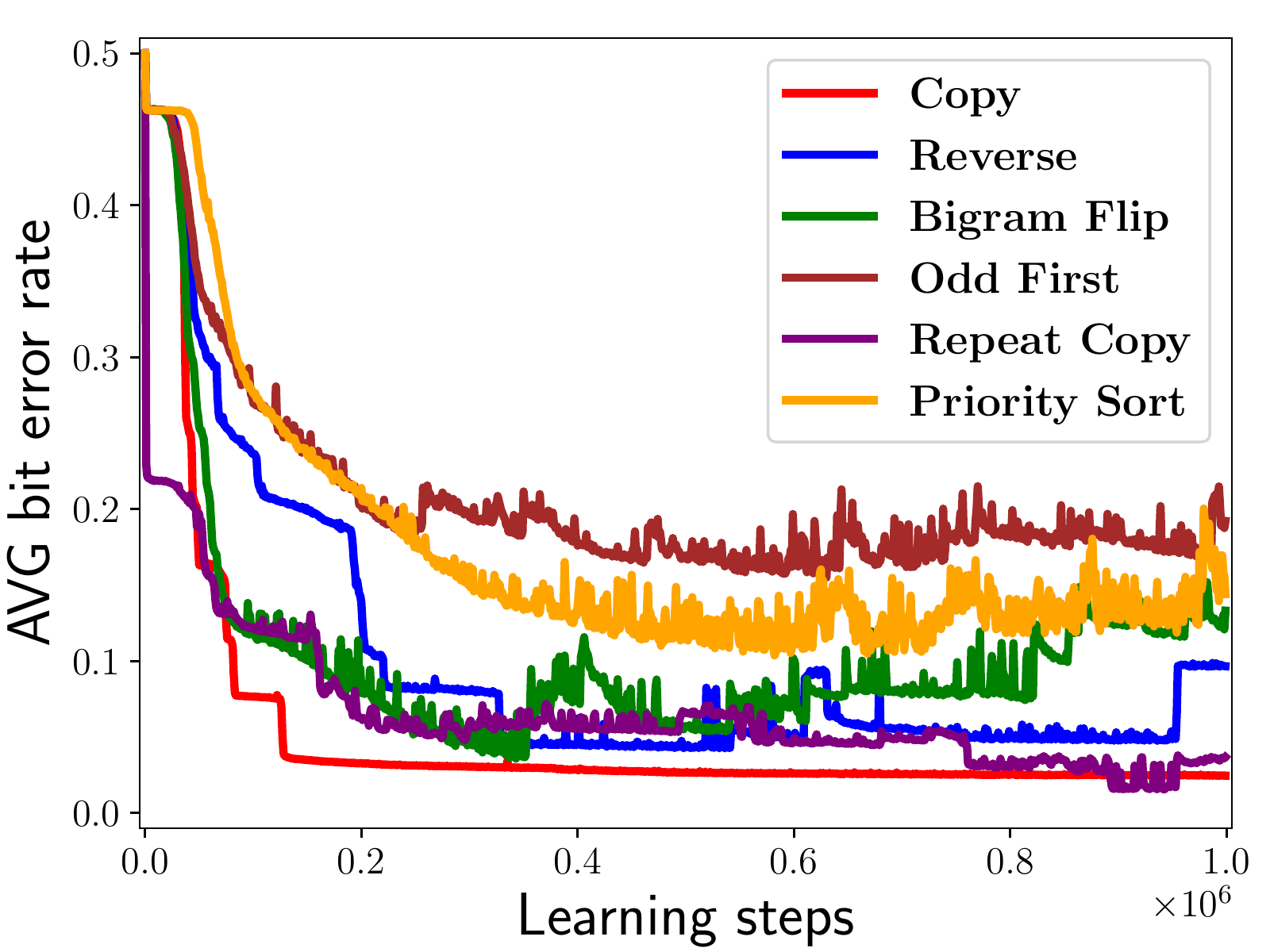} \label{dnc_dense_learning_curve}}
        \subfigure[NTM w/ EN controller]{\includegraphics[scale=0.25, bb=0 0 461 346]{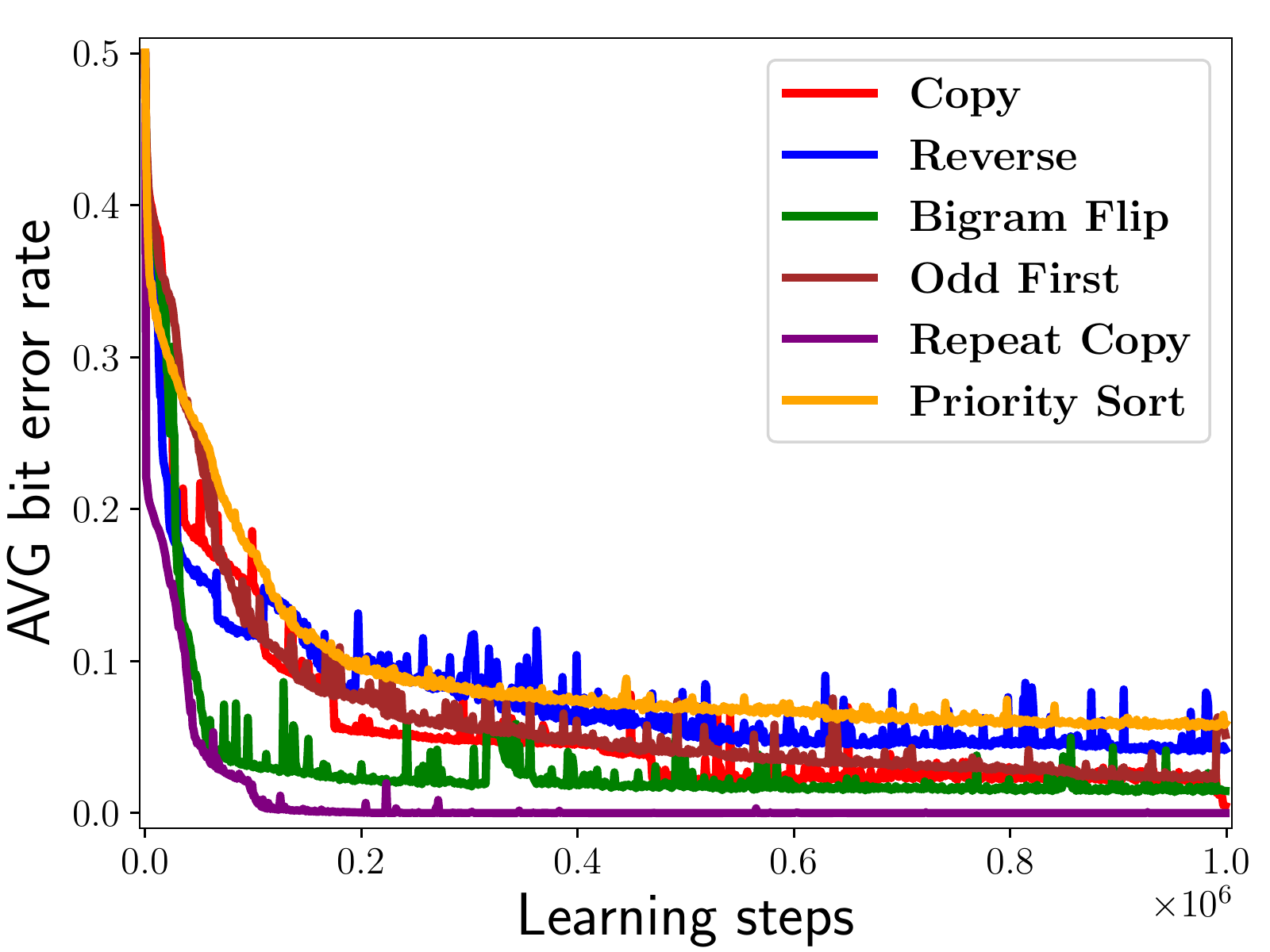} \label{ntm_recurrent_learning_curve}}
        \subfigure[DNC w/ EN controller]{\includegraphics[scale=0.25, bb=0 0 461 346]{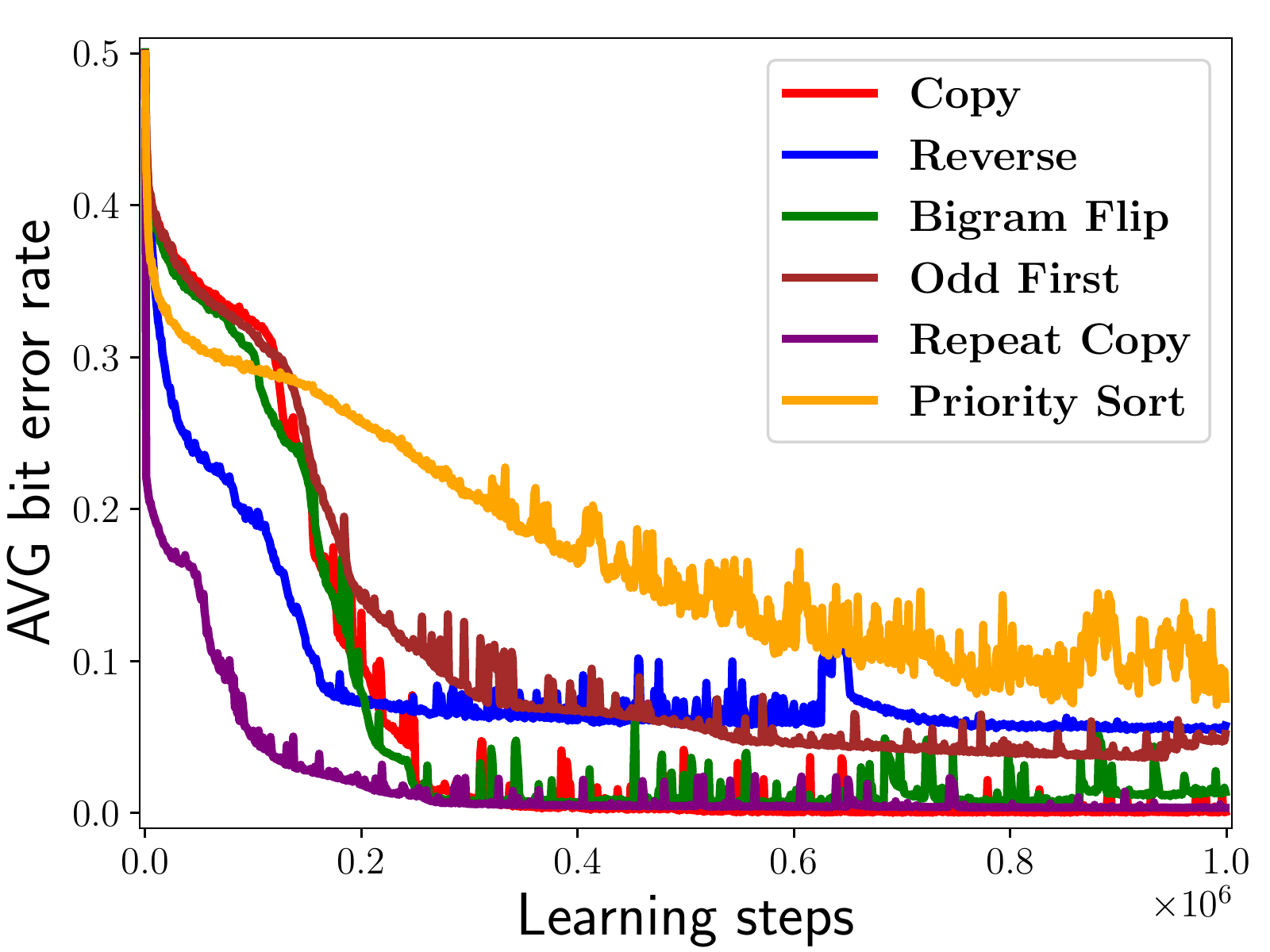} \label{dnc_recurrent_learning_curve}}
        \subfigure[NTM w/ proposed EN controller]{\includegraphics[scale=0.25, bb=0 0 461 346]{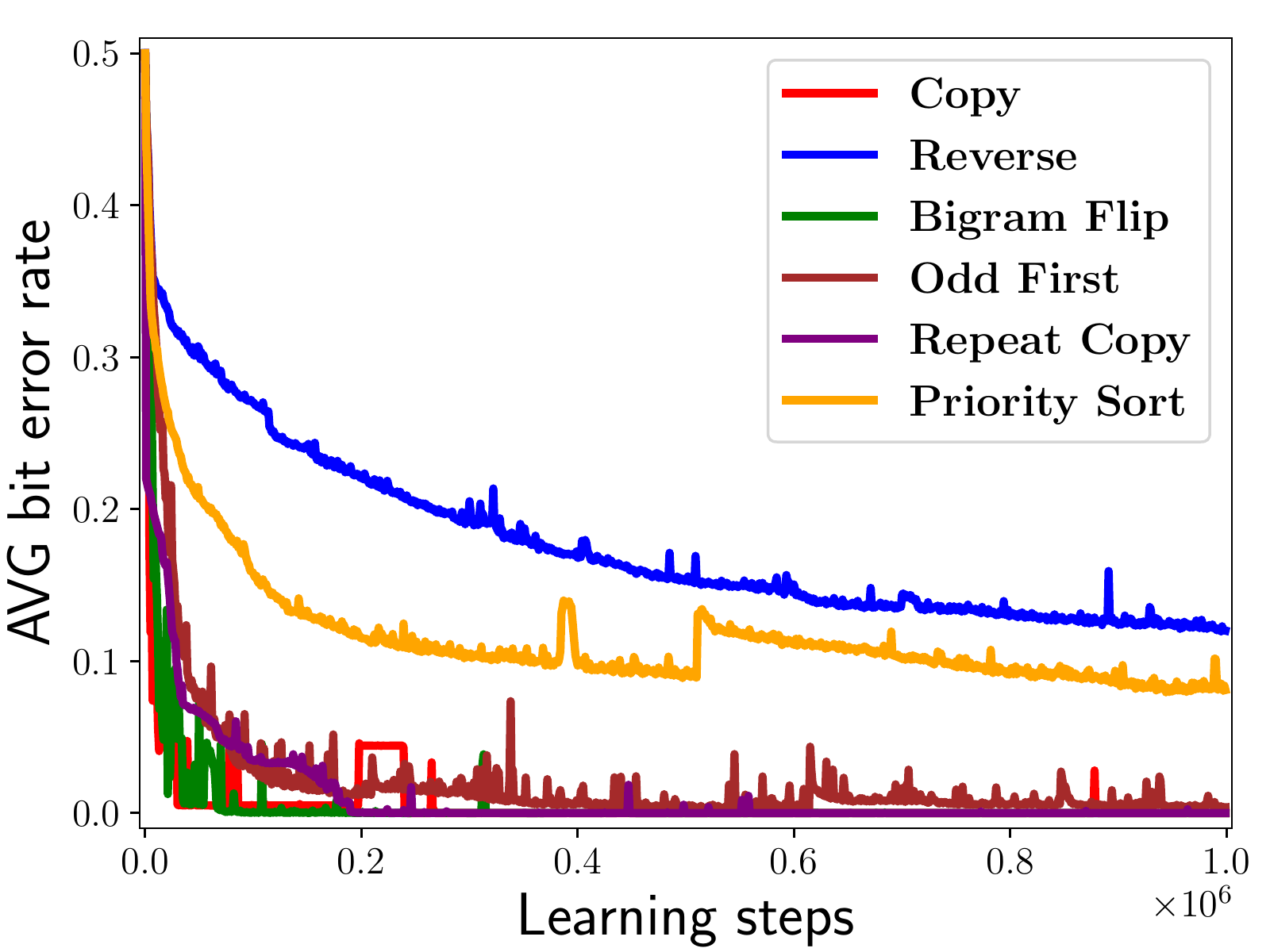} \label{ntm_proposed_recurrent_learning_curve}}
        \subfigure[DNC w/ proposed EN controller]{\includegraphics[scale=0.25, bb=0 0 461 346]{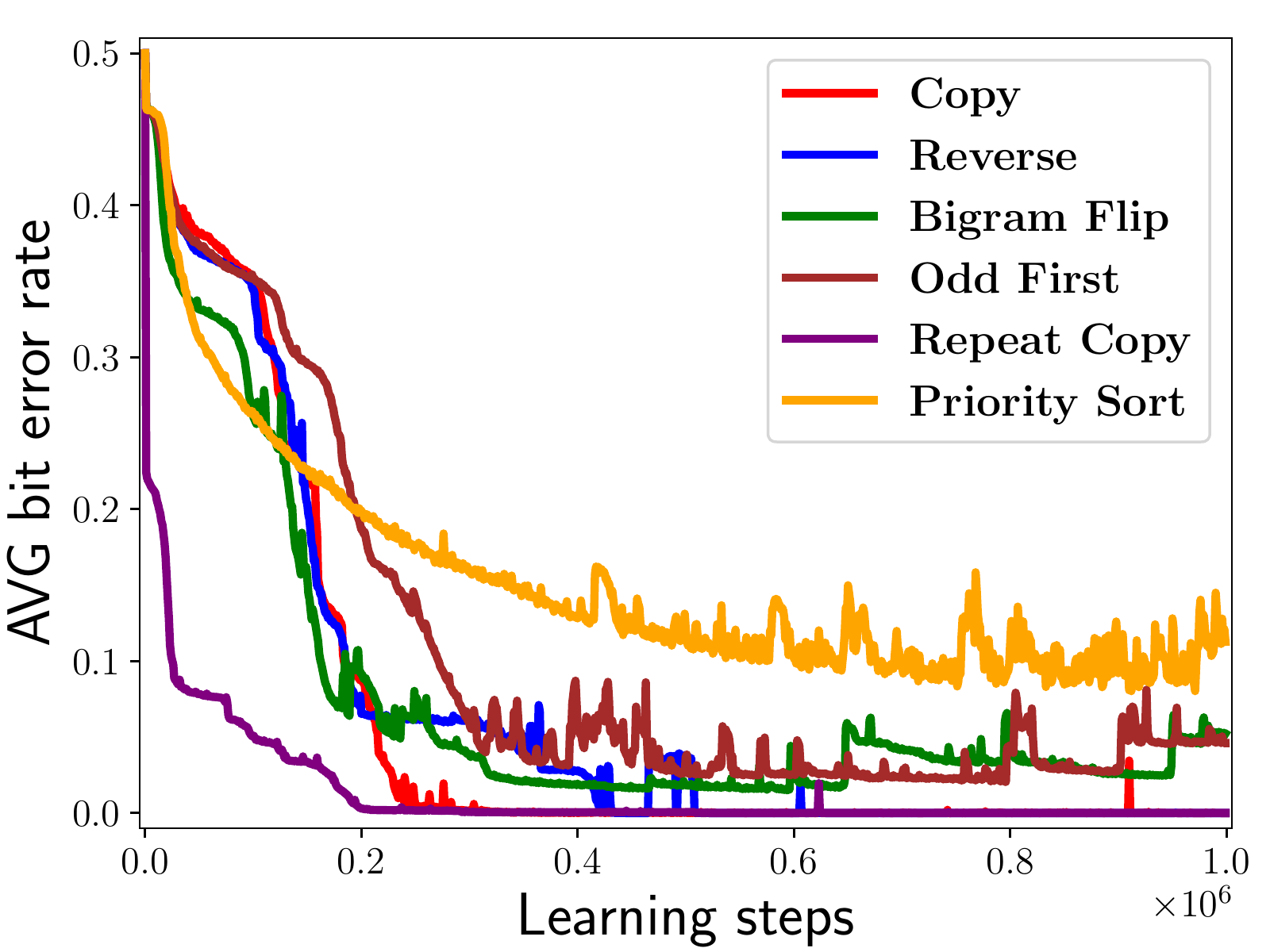} \label{dnc_proposed_recurrent_learning_curve}}
        \subfigure[NTM w/ LSTM controller]{\includegraphics[scale=0.25, bb=0 0 461 346]{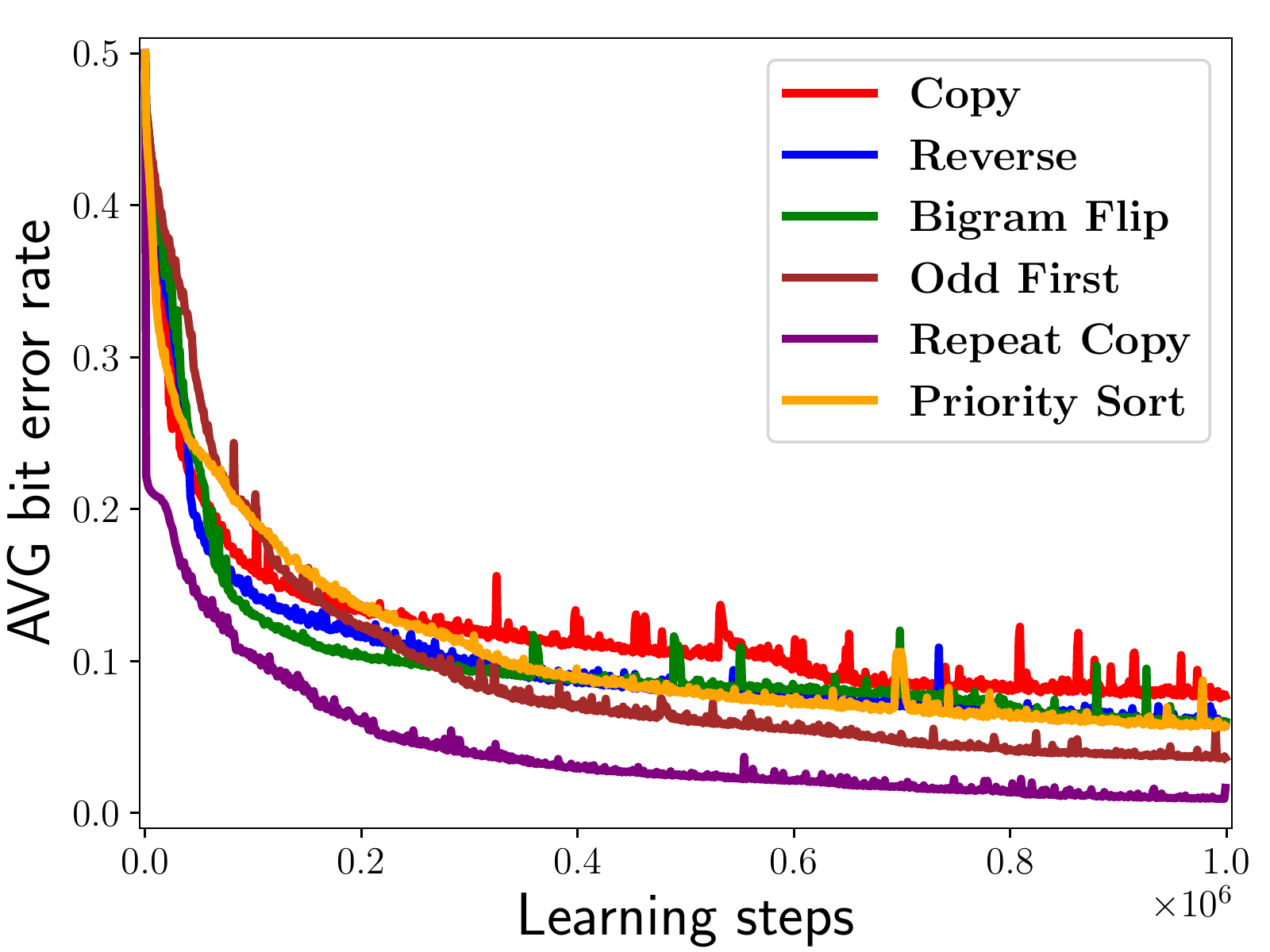} \label{ntm_lstm_learning_curve}}
        \subfigure[DNC w/ LSTM controller]{\includegraphics[scale=0.25, bb=0 0 461 346]{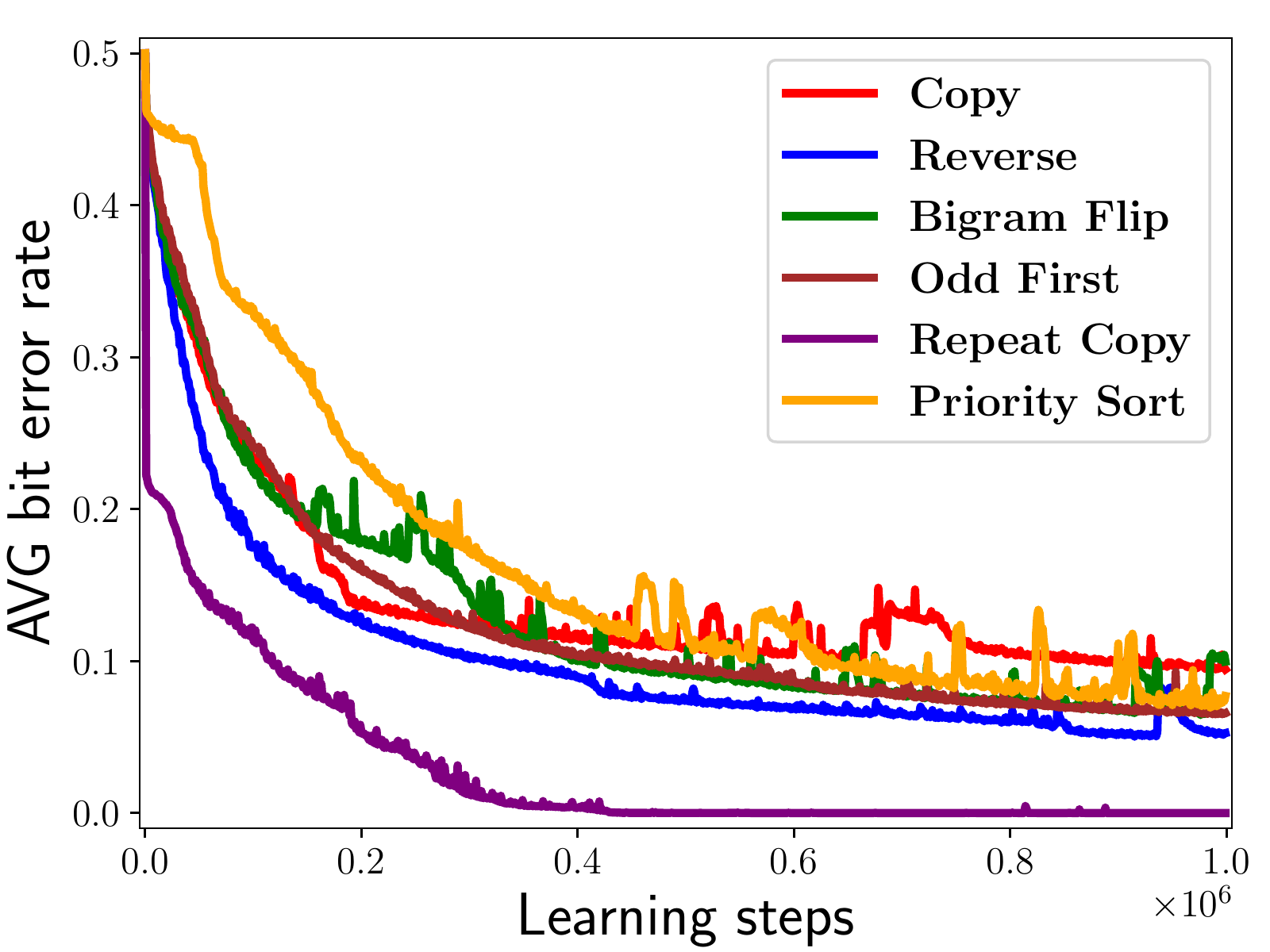} \label{dnc_lstm_learning_curve}}
        \subfigure[NTM w/ proposed LSTM controller]{\includegraphics[scale=0.25, bb=0 0 461 346]{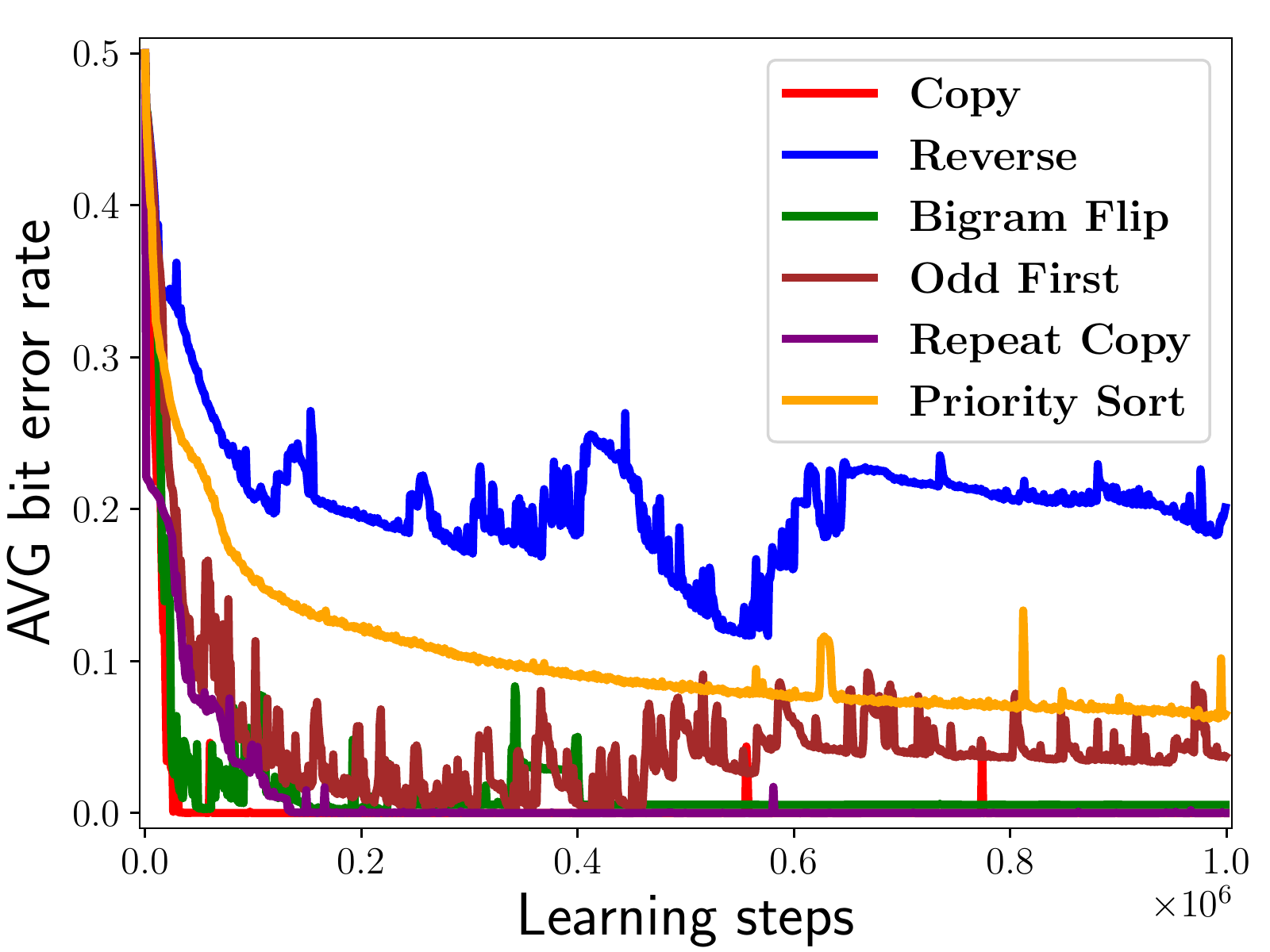} \label{ntm_proposed_lstm_learning_curve}}
        \subfigure[DNC w/ proposed LSTM controller]{\includegraphics[scale=0.25, bb=0 0 461 346]{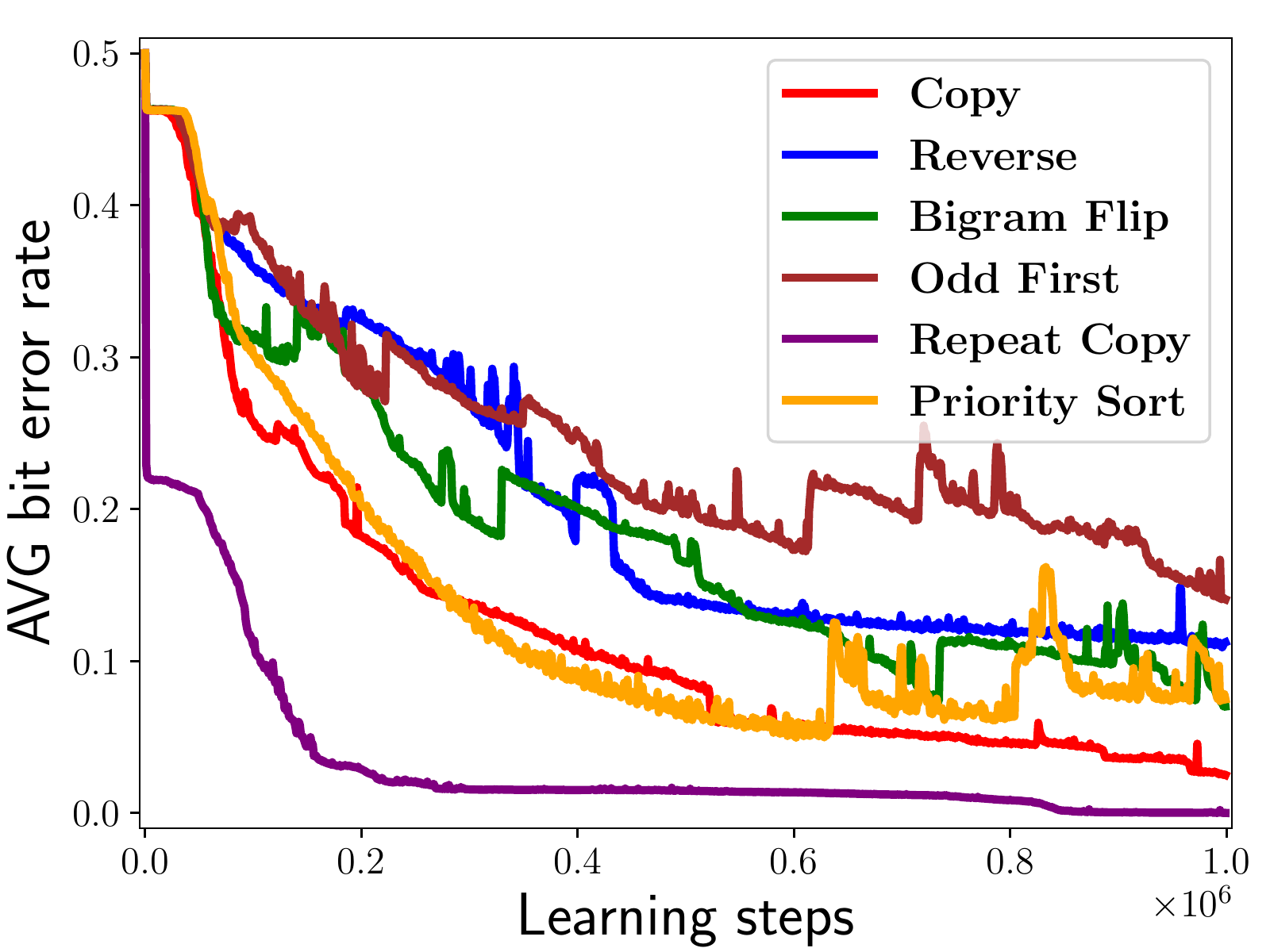} \label{dnc_proposed_lstm_learning_curve}}
  \caption{Average learning curves of the NTM and the DNC with all of the controllers we use. Note that the results reported in \tablename\ref{toy_result} were obtained at the best learning steps in terms of validation loss.}
  \label{learning_curves}
  \end{center}
\end{figure}

%\vspace{1mm}
\noindent
%\textbf{Discussions on the average learning curves.}
\textbf{Discussions on the average learning curves.}
\figurename\ref{learning_curves} shows the average learning curves of the NTM and the DNC with different controllers. In \figurename\ref{learning_curves}, we can see that the learning curves of successful case of the NTM converge so fast (e.g. the NTM with the FNN, the proposed EN, and the proposed LSTM controller on \textsc{Copy} or \textsc{Bigram Flip}). The architecture of the NTM is simple but contains enough functions to solve basic tasks such as \textsc{Copy}, while that of the DNC is more general but more complex. The FNN controller and the proposed RNN-based controllers utilize the external memory and functions of the model to solve the tasks as much as possible, which results in the fast convergence. 

On the other hand, the models with the FNN, the proposed EN, and the proposed LSTM controllers are less stable compared to those with the EN and the LSTM controllers. For example, the learning curve for \textsc{Reverse Copy} of the NTM with the proposed LSTM controller is basically worse than that of the NTM with the LSTM controller, while the test result with the proposed LSTM controller is better than that with the LSTM controller as seen in \tablename\ref{toy_result}. This is because the learning curves of the NTM with the proposed LSTM controller has high volatility. Therefore, using appropriate early stopping is important to achieve good performance on the models with the FNN or the proposed RNN-based controllers.

\subsection{bAbI Task}
\label{babi_task}
\textbf{Settings.} 
To evaluate the proposed controllers on more practical situations, we carry out experiments on the bAbI task, which is a set of $20$ simple question answering tasks. In each task, the models read stories followed by a few questions.
Because the experiments using the full dataset were too computationally expensive, we only used Tasks $1, \, 4, \, 9, \, 10, \, 11, \, 14$ of the $20$ task, following Hsin \shortcite{hsin_implementation_nodate}.
In the experiment, we train the models using a joint dataset of these six tasks, and the evaluation is conducted for each task separately.
We use the dataset provided by Facebook \footnote[1]{\url{http://www.thespermwhale.com/jaseweston/babi/tasks_1-20_v1-2.tar.gz}} with 10k training examples for each task. We use NNs with $128$ units for the controller, $128 \times 32$ for the memory, and $R=4$. The epoch size is $128$, and the other detailed settings are the same as Graves et al. \shortcite{dnc}.
%Following \cite{hsin_implementation_nodate}, we only use Tasks $1, \, 4, \, 9, \, 10, \, 11, \, 14$ of the $20$ tasks because experiments using the full dataset were too computationally expensive. We use the dataset provided by Facebook \footnote[2]{\url{http://www.thespermwhale.com/jaseweston/babi/tasks_1-20_v1-2.tar.gz}} with 10k training examples for each task. We use NNs with $128$ units for the controller, $128 \times 32$ for the memory , and $R=4$. The epoch size is $50$, and other detailed settings are the same as \cite{dnc}.

\vspace{1mm}
\noindent
\textbf{Discussions.} 
\tablename\ref{mean_babi_result} shows the average error rates of the ten models on the bAbI task. Due to the difference of the settings about the experiment, the results are different from those in Graves et al. \shortcite{dnc}. 

\tablename\ref{mean_babi_result} shows that the DNC with the proposed LSTM controller performs the best in terms of the mean error rate. The proposed LSTM controller brings out the potential ability of the DNC (e.g. the DNC can benefit from its design of tracking the order of written addresses to solve Task 14, time reasoning) because our proposed controller is designed to utilize the external memory. In addition, the proposed LSTM controller performs the best in all of the six tasks on each of the NTM and the DNC. Another observation seen in \tablename\ref{mean_babi_result} is that the scores of the models with the EN and the proposed EN controllers are worse than those of the other cases. We speculate the reason that the internal memory of the EN does not contribute to increasing the performance of the model while preventing it from converging to an appropriate solution. 

\section{Related Work}
\label{related_work}
%Among several types of studies about NTM-based MANNs, the most active one is about their architectures \cite{zaremba_reinforcement_2015,santoro_meta-learning_2016}. 
NTM-based MANNs have been actively studied since the advent of the NTM \cite{santoro_meta-learning_2016,park_quantized_2017,franke_robust_2018}. Rae et al. \shortcite{Rae2016ScalingMN} proposed Sparse Access Memory (SAM), which is a scalable end-to-end differentiable memory access scheme. One of the biggest restrictions of MANNs is that the capacity of memory depends on the size of the external memory, while larger external memory requires more computational cost. SAM enables efficient training of a MANN with a very large memory.
Zaremba and Sutskever \shortcite{zaremba_reinforcement_2015} used a reinforcement learning algorithm on the NTM to apply it for tasks that require discrete interfaces, which are not differentiable.
%NTM-based MANNs have been actively studied since the advent of the NTM \cite{zaremba_reinforcement_2015,Rae2016ScalingMN,santoro_meta-learning_2016}. 
%Although they have been proposed only recently, NTM-based MANNs are actively studied \cite{zaremba_reinforcement_2015,Rae2016ScalingMN,santoro_meta-learning_2016}. 
%Although the focus of most of those studies is the refinement of the read heads, the write head, and the memory, some studies refine controllers.
%Although those studies focus on various themes, few ones address refinements on the controller.

Gulcehre et al. \shortcite{dntm} proposed a novel NTM-based MANN, Dynamic Neural Turing Machine (D-NTM). While the original NTM implements location-based addressing using shift operations with a fixed size, the D-NTM performs this operation using NNs directly. They also address the same problem as ours, but they tackle the problem by using a regularization approach where the addresses pointed by the read heads and the write head are forced to be consistent. 
Gulcehre et al. \shortcite{gulcehre_memory_2017} proposed a novel MANN called TARDIS based on the concept that MANNs connect ``time'' discontinuously. Their work also addresses the problem which we focus on, but their approach is using $\bm{\mathcal{H}}_t^o = [\bm{x}_t \, ; \, \bm{r}_{t}]$ for the output of the model. Our proposed controller uses $\bm{h}'_t$ instead of $\bm{x}_t$ for $\bm{\mathcal{H}}_t^o$, which enables the output of the model and written information at $t$ to be more consistent.
%\textbf{This is similar to ours, but our proposal requires no extra parameters to train.}

There also exist models called MANNs which are not based on the NTM. In particular, the models based on Memory Networks \cite{weston_memory_2014} are actively studied \cite{kumar_ask_2015,sukhbaatar_end--end_2015,entity_net}. These MANNs are different from the NTM-based models in some respects (e.g. MANNs based on Memory Networks conduct the read operation multiple times in a time step). In addition, MANNs based on Memory Networks do not have a module corresponding to the controller, which controls all the other modules.

\section{Conclusion and Future Work}
\label{conclusion}
In this paper, we have proposed a novel type of RNN-based controller for MANNs. Without increasing the number of training parameters, our proposed controller avoids using the memory in the controller for the output of the model and benefits from it for controlling the other modules. 
In the experiments on both of the Toy and bAbI tasks, the best scores are achieved by the  models with our proposed controller, which demonstrates the effectiveness of our approach. 

An interesting direction of future work is exploring other architectures based on the insights obtained in this work because there are more variations than the two proposed controllers we have used.

\bibliographystyle{named}
%\bibliography{ijcai18}
%\bibliographystyle{jplain}
%{\normalsize 
%\input{arxiv1st.bbl}
% \bibliography{bib_ref}

\begin{thebibliography}{}

\bibitem[\protect\citeauthoryear{Cho \bgroup \em et al.\egroup
  }{2014a}]{enc_dec}
Kyunghyun Cho, B~{van Merrienboer}, Dzmitry Bahdanau, and Yoshua Bengio.
\newblock On the properties of neural machine translation: Encoder-decoder
  approaches.
\newblock {\em Eighth Workshop on Syntax, Semantics and Structure in
  Statistical Translation}, 2014.

\bibitem[\protect\citeauthoryear{Cho \bgroup \em et al.\egroup
  }{2014b}]{cho_learning_2014}
Kyunghyun Cho, Bart van Merrienboer, Caglar Gulcehre, Dzmitry Bahdanau, Fethi
  Bougares, Holger Schwenk, and Yoshua Bengio.
\newblock Learning {Phrase} {Representations} using {RNN}
  {Encoder}^^e2^^80^^93{Decoder} for {Statistical} {Machine} {Translation}.
\newblock In {\em EMNLP}, 2014.

\bibitem[\protect\citeauthoryear{Elman}{1990}]{elman1990finding}
Jeffrey~L Elman.
\newblock Finding structure in time.
\newblock {\em Cognitive science}, 1990.

\bibitem[\protect\citeauthoryear{Franke \bgroup \em et al.\egroup
  }{2018}]{franke_robust_2018}
J^^c3^^b6rg Franke, Jan Niehues, and Alex Waibel.
\newblock Robust and {Scalable} {Differentiable} {Neural} {Computer} for
  {Question} {Answering}.
\newblock {\em arXiv:1807.02658 [cs]}, 2018.

\bibitem[\protect\citeauthoryear{Graves \bgroup \em et al.\egroup
  }{2013}]{Graves2013SpeechRW}
Alex Graves, Abdel rahman Mohamed, and Geoffrey~E. Hinton.
\newblock Speech recognition with deep recurrent neural networks.
\newblock {\em IEEE ICASSP}, 2013.

\bibitem[\protect\citeauthoryear{Graves \bgroup \em et al.\egroup
  }{2014}]{graves_neural_2014}
Alex Graves, Greg Wayne, and Ivo Danihelka.
\newblock Neural turing machines.
\newblock {\em arXiv preprint arXiv:1410.5401}, 2014.

\bibitem[\protect\citeauthoryear{Graves \bgroup \em et al.\egroup }{2016}]{dnc}
Alex Graves, Greg Wayne, Malcolm Reynolds, Tim Harley, Ivo Danihelka, Agnieszka
  Grabska-Barwi^^c5^^84ska, Sergio~G^^c3^^b3mez Colmenarejo, Edward
  Grefenstette, Tiago Ramalho, John Agapiou, and {others}.
\newblock Hybrid computing using a neural network with dynamic external memory.
\newblock {\em Nature}, 2016.

\bibitem[\protect\citeauthoryear{Graves}{2013}]{graves2013generating}
Alex Graves.
\newblock Generating sequences with recurrent neural networks.
\newblock {\em arXiv preprint arXiv:1308.0850}, 2013.

\bibitem[\protect\citeauthoryear{Grefenstette \bgroup \em et al.\egroup
  }{2015}]{grefenstette_learning_2015}
Edward Grefenstette, Karl~Moritz Hermann, Mustafa Suleyman, and Phil Blunsom.
\newblock Learning to transduce with unbounded memory.
\newblock In {\em NIPS}, 2015.

\bibitem[\protect\citeauthoryear{Gulcehre \bgroup \em et al.\egroup
  }{2017a}]{gulcehre_memory_2017}
Caglar Gulcehre, Sarath Chandar, and Yoshua Bengio.
\newblock Memory {Augmented} {Neural} {Networks} with {Wormhole} {Connections}.
\newblock {\em arXiv:1701.08718}, 2017.

\bibitem[\protect\citeauthoryear{Gulcehre \bgroup \em et al.\egroup
  }{2017b}]{dntm}
Caglar Gulcehre, Sarath Chandar, Kyunghyun Cho, and Yoshua Bengio.
\newblock Dynamic {Neural} {Turing} {Machine} with {Soft} and {Hard}
  {Addressing} {Schemes}.
\newblock {\em arXiv:1607.00036}, 2017.

\bibitem[\protect\citeauthoryear{Henaff \bgroup \em et al.\egroup
  }{2016}]{entity_net}
Mikael Henaff, Jason Weston, Arthur Szlam, Antoine Bordes, and Yann LeCun.
\newblock Tracking the world state with recurrent entity networks.
\newblock {\em arXiv:1612.03969}, 2016.

\bibitem[\protect\citeauthoryear{Hochreiter and
  Schmidhuber}{1997}]{hochreiter1997long}
Sepp Hochreiter and J{\"u}rgen Schmidhuber.
\newblock Long short-term memory.
\newblock {\em Neural computation}, 1997.

\bibitem[\protect\citeauthoryear{Hsin}{2017}]{hsin_implementation_nodate}
Carol Hsin.
\newblock Implementation and {Optimization} of {Differentiable} {Neural}
  {Computers}.
\newblock {\em Technical Report for CS224 at Stanford University}, 2017.

\bibitem[\protect\citeauthoryear{Kumar \bgroup \em et al.\egroup
  }{2015}]{kumar_ask_2015}
Ankit Kumar, Ozan Irsoy, Jonathan Su, James Bradbury, Robert English, Brian
  Pierce, Peter Ondruska, Ishaan Gulrajani, and Richard Socher.
\newblock Ask me anything: {Dynamic} memory networks for natural language
  processing.
\newblock {\em arXiv preprint arXiv:1506.07285}, 2015.

\bibitem[\protect\citeauthoryear{Oord \bgroup \em et al.\egroup
  }{2016}]{oord_wavenet}
Aaron van~den Oord, Sander Dieleman, Heiga Zen, Karen Simonyan, Oriol Vinyals,
  Alex Graves, Nal Kalchbrenner, Andrew Senior, and Koray Kavukcuoglu.
\newblock {WaveNet}: {A} {Generative} {Model} for {Raw} {Audio}.
\newblock {\em arXiv:1609.03499}, 2016.

\bibitem[\protect\citeauthoryear{Park \bgroup \em et al.\egroup
  }{2017}]{park_quantized_2017}
Seongsik Park, Seijoon Kim, Seil Lee, Ho~Bae, and Sungroh Yoon.
\newblock Quantized {Memory}-{Augmented} {Neural} {Networks}.
\newblock {\em arXiv:1711.03712 [cs, stat]}, 2017.

\bibitem[\protect\citeauthoryear{Rae \bgroup \em et al.\egroup
  }{2016}]{Rae2016ScalingMN}
Jack~W. Rae, Jonathan~J. Hunt, Ivo Danihelka, Timothy Harley, Andrew~W. Senior,
  Gregory Wayne, Alex Graves, and Tim Lillicrap.
\newblock Scaling memory-augmented neural networks with sparse reads and
  writes.
\newblock In {\em NIPS}, 2016.

\bibitem[\protect\citeauthoryear{Santoro \bgroup \em et al.\egroup
  }{2016}]{santoro_meta-learning_2016}
Adam Santoro, Sergey Bartunov, Matthew Botvinick, Daan Wierstra, and Timothy
  Lillicrap.
\newblock Meta-learning with memory-augmented neural networks.
\newblock In {\em ICML}, 2016.

\bibitem[\protect\citeauthoryear{Sukhbaatar \bgroup \em et al.\egroup
  }{2015}]{sukhbaatar_end--end_2015}
Sainbayar Sukhbaatar, Jason Weston, Rob Fergus, and {others}.
\newblock End-to-end memory networks.
\newblock In {\em NIPS}, 2015.

\bibitem[\protect\citeauthoryear{Weston \bgroup \em et al.\egroup
  }{2014}]{weston_memory_2014}
Jason Weston, Sumit Chopra, and Antoine Bordes.
\newblock Memory networks.
\newblock {\em arXiv preprint arXiv:1410.3916}, 2014.

\bibitem[\protect\citeauthoryear{Weston \bgroup \em et al.\egroup
  }{2015}]{weston2015towards}
Jason Weston, Antoine Bordes, Sumit Chopra, Alexander~M Rush, Bart van
  Merri{\"e}nboer, Armand Joulin, and Tomas Mikolov.
\newblock Towards ai-complete question answering: A set of prerequisite toy
  tasks.
\newblock {\em arXiv preprint arXiv:1502.05698}, 2015.

\bibitem[\protect\citeauthoryear{Wu \bgroup \em et al.\egroup
  }{2016}]{wu_googles_2016}
Yonghui Wu, Mike Schuster, Zhifeng Chen, Quoc~V. Le, Mohammad Norouzi, Wolfgang
  Macherey, Maxim Krikun, Yuan Cao, Qin Gao, Klaus Macherey, Jeff Klingner,
  Apurva Shah, Melvin Johnson, Xiaobing Liu, ^^c5^^81ukasz Kaiser, Stephan
  Gouws, Yoshikiyo Kato, Taku Kudo, Hideto Kazawa, Keith Stevens, George
  Kurian, Nishant Patil, Wei Wang, Cliff Young, Jason Smith, Jason Riesa, Alex
  Rudnick, Oriol Vinyals, Greg Corrado, Macduff Hughes, and Jeffrey Dean.
\newblock Google's {Neural} {Machine} {Translation} {System}: {Bridging} the
  {Gap} between {Human} and {Machine} {Translation}.
\newblock {\em arXiv:1609.08144}, 2016.

\bibitem[\protect\citeauthoryear{Yang and Rush}{2016}]{yang_lie-access_2016}
Greg Yang and Alexander~M. Rush.
\newblock Lie-{Access} {Neural} {Turing} {Machines}.
\newblock {\em arXiv:1611.02854}, 2016.

\bibitem[\protect\citeauthoryear{Zaremba and
  Sutskever}{2015}]{zaremba_reinforcement_2015}
Wojciech Zaremba and Ilya Sutskever.
\newblock Reinforcement {Learning} {Neural} {Turing} {Machines} - {Revised}.
\newblock {\em arXiv:1505.00521}, 2015.

\end{thebibliography}
%}

\end{document}